\documentclass[10pt,twocolumn,letterpaper]{article}

\usepackage{iccv}
\usepackage{times}
\usepackage{epsfig}
\usepackage{graphicx}
\usepackage{amsmath}
\usepackage{amssymb}
\usepackage{floatrow}
\usepackage{adjustbox}
\usepackage{booktabs}
\usepackage[utf8]{inputenc} %
\usepackage[T1]{fontenc}    %
\usepackage{url}            %
\usepackage{amsfonts}       %
\usepackage{stfloats}
\usepackage[accsupp]{axessibility}

\usepackage{nicefrac}       %
\usepackage{microtype}      %
\usepackage{mwe}
\usepackage{wrapfig}

\usepackage{color, colortbl}
\usepackage{algorithm}
\usepackage[noend]{algpseudocode}
\usepackage{pifont}
\usepackage{paralist}
\usepackage{multirow}
\usepackage{xcolor}
\usepackage{xspace}
\usepackage{subcaption}
\usepackage[font=small]{caption}
\usepackage{mwe}
\usepackage{comment}
\usepackage[numbers,sort,compress]{natbib}
\usepackage{etoc}

\newfloatcommand{capbtabbox}{table}
\newfloatcommand{capbfigbox}{figure}

\definecolor{Gray}{gray}{0.9}

\usepackage[pagebackref=true,breaklinks=true,colorlinks,bookmarks=false]{hyperref}

\newcommand{\method}{\texttt{SENTRY}\xspace}
\newcommand{\source}{\mathcal{S}}  %
\newcommand{\target}{\mathcal{T}}  %

\newcommand{\numclasses}{C}  %
\newcommand{\model}{h}  %
\newcommand{\modelparams}{\Theta}  %

\DeclareMathOperator*{\argmin}{argmin}
\DeclareMathOperator*{\argmax}{argmax}

\newcommand{\veryshortarrow}[1][3pt]{\mathrel{%
   \hbox{\rule[\dimexpr\fontdimen22\textfont2-.2pt\relax]{#1}{.4pt}}%
   \mkern-4mu\hbox{\usefont{U}{lasy}{m}{n}\symbol{41}}}}

\usepackage[breaklinks=true,bookmarks=false]{hyperref}

\iccvfinalcopy %

\def\iccvPaperID{1713} %

\begin{document}

\title{SENTRY: Selective Entropy Optimization via Committee Consistency for Unsupervised Domain Adaptation}

\author{
    \textbf{Viraj Prabhu}\qquad
    \textbf{Shivam Khare} \qquad
    \textbf{Deeksha Kartik} \qquad 
    \textbf{Judy Hoffman} \qquad \\
    Georgia Institute of Technology \\
    {\small\texttt{\{virajp,skhare31,dkartik3,judy\}@gatech.edu}}
}

\maketitle
\ificcvfinal\thispagestyle{empty}\fi

\vspace{-25pt}
\begin{abstract}
    \vspace{-7pt}
   Many existing approaches for unsupervised domain adaptation (UDA) focus on adapting under only data distribution shift and offer limited success under additional cross-domain label distribution shift. Recent work based on self-training using target pseudolabels has shown promise, but on challenging shifts pseudolabels may be highly unreliable and using them for self-training may lead to error accumulation and domain misalignment. We propose Selective Entropy Optimization via Committee Consistency (\method), a UDA algorithm that judges the reliability of a target instance based on its predictive consistency under a committee of random image transformations. Our algorithm then selectively minimizes predictive entropy to increase confidence on highly consistent target instances, while maximizing predictive entropy to reduce confidence on highly inconsistent ones. In combination with pseudolabel-based approximate target class balancing, our approach leads to  significant improvements over the state-of-the-art on 27/31 domain shifts from standard UDA benchmarks as well as benchmarks designed to stress-test adaptation under label distribution shift. Our code is available at \url{https://github.com/virajprabhu/SENTRY}.
    \vspace{-7pt}
\end{abstract}
\vspace{-15pt}

\section{Introduction}
\label{ref:intro}
\vspace{-5pt}

Unsupervised domain adaptation (UDA) learns to transfer a predictive model from a labeled source domain to an unlabeled target domain. The particular instantiation of learning under covariate shift has been extensively studied within the computer vision community~\cite{ganin2014unsupervised,hoffman2017cycada,long2015learning,saenko2010adapting,tzeng2017adversarial,tzeng2014deep}.
However, many modern UDA methods, such as distribution matching based techniques,
implicitly assume that the task label distribution does not change across domains, i.e $P_\mathcal{S}(y) = P_\mathcal{T}(y)$. When such an assumption is violated, distribution matching is not expected to succeed~\cite{li2020rethinking,wu2019domain}.
   
In many real-world adaptation scenarios, one may encounter data distribution (\ie covariate) shift across domains together with label distribution shift (LDS). For instance, a source dataset can be curated to have a balanced label distribution while a naturally arising target dataset may follow a power law label distribution, as some categories naturally occur more often than others (\eg DomainNet~\cite{peng2019moment}, LVIS~\cite{gupta2019lvis}, and MSCOCO~\cite{lin2014microsoft}). In order to make domain adaptation broadly applicable, it is critical to develop UDA algorithms that can operate under joint data and label distribution shift.
   
\begin{figure}[t]
   \centering
    \vspace{-10pt}
   \noindent\includegraphics[width=\linewidth]{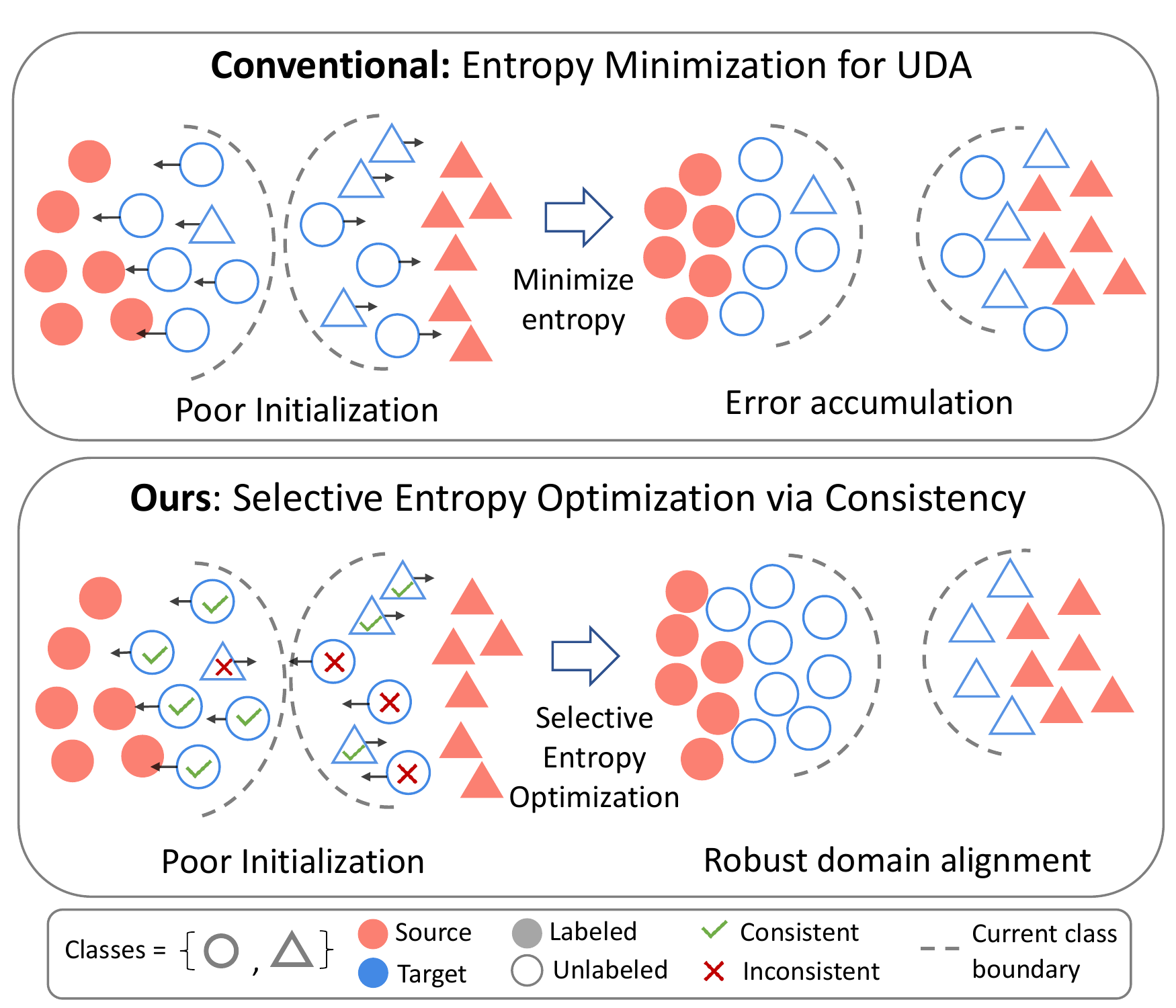}
   \caption{
   \textbf{Top}: Conventional entropy-minimization based approaches for unsupervised domain adaptation (UDA) operate by increasing model confidence on unlabeled target instances. Under strong distribution shifts, some instances may initially be misaligned and entropy minimization can lead to error accumulation. \textbf{Bottom}: We propose Selective Entropy Optimization via Committee Consistency (\method), a UDA algorithm that i) identifies reliable target instances based on their predictive consistency under a set of random image transformations, and ii) selectively optimizes model entropy on these instances to induce domain alignment. %
   }
   \label{fig:teaser}
   \vspace*{-7pt}
\end{figure}

Recent works have attempted to address the problem of joint data and label distribution shift~\cite{li2020rethinking,tan2019generalized}, but these approaches can be unstable as they rely on self-training using often noisy pseudo-labels or conditional entropy minimization~\cite{li2020rethinking} over potentially miscalibrated predictions~\cite{guo2017calibration,snoek2019can}. Thus, when learning with unconstrained self-training, early mistakes can result in error accumulation~\cite{chen2019progressive} and significant domain misalignment (see Figure~\ref{fig:teaser}, top).

To address the problem of error accumulation arising from unconstrained self-training, we propose Selective Entropy Optimization via Committee Consistency (\method), a novel \emph{selective} self-training algorithm for UDA. First, rather than using model confidence which can be miscalibrated under a domain shift~\cite{snoek2019can}, we identify reliable target instances for self-training based on their \emph{predictive consistency} under a committee of random, label-preserving image transformations. Such consistency checks have been found to be a reliable way to detect model errors~\cite{bahat2019natural}. Having identified reliable and unreliable target instances, we then perform selective entropy optimization: we consider a highly consistent target instance as likely correctly aligned, and increase model confidence by minimizing predictive entropy for such an instance. Similarly, we consider an instance with high predictive inconsistency over transformations as likely misaligned, and reduce model confidence by \emph{maximizing} predictive entropy. See Figure~\ref{fig:teaser} (bottom).

\noindent \textbf{Contributions.} We propose \method, an algorithm for unsupervised adaptation under simultaneous data and label distribution shift. We make the following contributions:
\begin{compactenum}
   \item A novel selection criterion that identifies reliable target instances for self-training based on predictive consistency over a committee of random, label-preserving image transformations.
   \item A selective entropy optimization objective that minimizes predictive entropy (increasing confidence) on highly consistent target instances, and maximizes it (reducing confidence) on highly inconsistent ones.
   \item We propose using class-balanced sampling on the source (using labels) and target (using pseudolabels), and find it to complement adaptation under LDS.
   \item \method sets a new state-of-the-art on 27/31 domain shifts belonging to both standard and LDS versions of several DA benchmarks for classification, including DomainNet~\cite{peng2019moment}, OfficeHome~\cite{venkateswara2017deep}, and VisDA~\cite{peng2017visda}.
\end{compactenum}

\vspace{-5pt}
\section{Related Work}
\label{ref:relwork}
\vspace{-4pt}

\noindent\textbf{Unsupervised Domain Adaptation (UDA)}. The task of transferring models from a labeled source to an unlabeled target domain has seen considerable progress~\cite{ganin2014unsupervised,hoffman2017cycada,saenko2010adapting,tzeng2014deep}. Many approaches align feature spaces via directly minimizing domain discrepancy statistics~\cite{kang2019contrastive,long2015learning,tzeng2014deep}. Recently, distribution matching (DM) via domain-adversarial learning has become a prominent UDA paradigm~\cite{ganin2014unsupervised,long2018conditional,saito2018adversarial,tzeng2017adversarial,zhang2019bridging}. Such DM-based methods however achieve limited success in the presence of additional label distribution shift (LDS). 

Some prior work has studied the problem of UDA under LDS, proposing class-weighted domain discrepancy measures~\cite{wang2017balanced,yan2017mind}, generative approaches for pair-wise feature matching~\cite{takahashi2020partially}, or asymmetrically-relaxed distribution alignment~\cite{you2019universal}. Some prior work in UDA under LDS additionally assumes that the conditional input distribution does not change across domains i.e. $p_\source(y)\neq p_\target(y), p_\source(x|y)=p_\target(x|y)$ (referred to as ``label shift'' ~\cite{azizzadenesheli2018regularized,lipton2018detecting,tachet2020domain}). We tackle the problem of UDA under simultaneous covariate and label distribution shift, without making additional assumptions.

\noindent \textbf{Self-training for UDA.} Recently, training on model predictions or \emph{self-training} has proved to be a promising approach for UDA under LDS~\cite{tan2019generalized,li2020rethinking}. This typically involves supervised training on confidently predicted target pseudolabels~\cite{tan2019generalized}, confidence regularization~\cite{zou2019confidence}, or conditional entropy minimization~\cite{grandvalet2005semi} on target instances~\cite{li2020rethinking}. %
However, unconstrained self-training can lead to error accumulation. To address this, we propose a \emph{selective} self-training strategy that first identifies reliable instances for self-training and selectively optimizes model entropy on those.

\noindent\textbf{Predictive Consistency.} Predictive consistency under augmentations has been found to be useful in several capacities -- as a regularizer in supervised learning~\cite{cubuk2020randaugment}, self-supervised representation learning~\cite{chen2020simple}, semi-supervised learning~\cite{sajjadi2016regularization,berthelot2019mixmatch,xie2020unsupervised,sohn2020fixmatch}, and UDA~\cite{li2020rethinking}. Bahat ~\etal~\cite{bahat2019natural} find consistency under image transformations to be a reliable indicator of model errors.
Unlike prior work which optimizes for invariance across augmentations, we use predictive consistency under a committee of random image transforms to \emph{detect} reliable instances for alignment, and selectively optimize model entropy on such instances.

\vspace{-5pt}
\section{Approach}
\label{sec:approach}
\vspace{-3pt}

We address the problem of unsupervised domain adaptation (UDA) of a model trained on a labeled source domain to an unlabeled target domain. In addition to covariate shift across domains, we focus on the practical scenario of additional cross-domain label distribution shift (LDS), and present a selective self-training algorithm for UDA that leads to reliable domain alignment in such a setting.

\vspace{-2pt}
\subsection{Notation}
\vspace{-2pt}

\noindent Let $\mathcal{X}$ and $\mathcal{Y}$ denote input and ouput spaces, with the goal being to learn a CNN mapping $\model: \mathcal{X} \to \mathcal{Y}$ parameterized by $\modelparams$. In unsupervised DA we are given access to labeled source instances $(\mathbf{x}_\source, y_\source) \sim \mathcal{P}_\source(\mathcal{X}, \mathcal{Y})$, and unlabeled target instances $\mathbf{x}_\target \sim \mathcal{P}_\target(\mathcal{X})$, where $\source$ and $\target$ correspond to source and target domains. We consider DA in the context of $C$-way image classification: the inputs $\mathbf{x}$ are images, and labels $y$ are categorical variables $y \in \{1, 2, .. , \numclasses \}$. 
For an instance $\mathbf{x}$, let $p_{\modelparams}(y|\mathbf{x})$ denote the final probabilistic output from the model. For each target instance $\mathbf{x}_\target \sim \mathcal{P}_\target(\mathcal{X})$, we estimate a pseudolabel ${\hat{y}} = \argmax p_{\modelparams}(y | \mathbf{x}_\target)$.

\vspace{-2pt}
\subsection{Preliminaries: UDA via entropy minimization}
\vspace{-2pt}

Unsupervised domain adaptation typically follows a two-stage training pipeline: source training, followed by target adaptation. In the first stage, a model is trained on the labeled source domain in a supervised fashion, minimizing a cross-entropy loss with respect to ground truth labels.

\begin{equation}
    \vspace{-7pt}
    \mathcal{L}_{CE} = \mathbb{E}_{(\mathbf{x}_\source, y_\source) \sim \mathcal{P}_\source} [\mathcal{L}_{CE} (\model(\mathbf{x}_\source), y_\source)]    
    \label{eq:ce}
\end{equation}    
\vspace{-5pt}

\begin{figure*}[t]
    \centering
    \vspace{-2pt}
    \includegraphics[width=\linewidth]{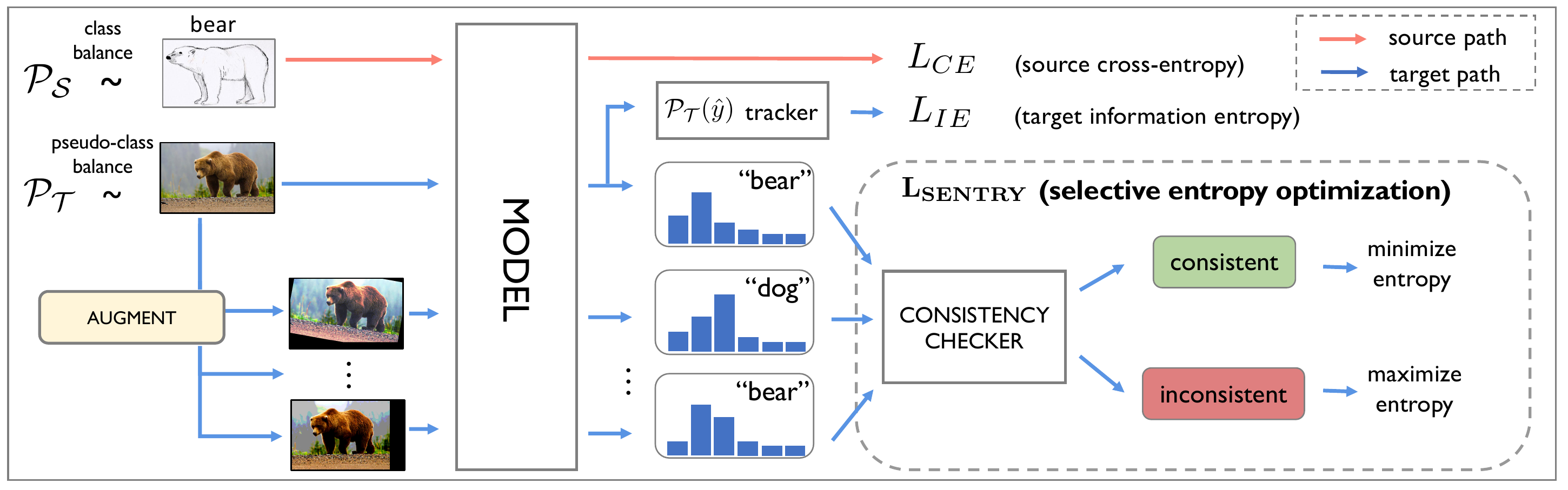}
    \vspace{-12pt}
    \caption{
    We propose Selective Entropy Optimization via Committee Consistency (\method) for unsupervised DA. For each target instance, we generate a committee of random, label-preserving image transformations. A consistency checker then computes the consistency between model predictions for the original and augmented versions. The algorithm then minimizes predictive entropy (increasing model confidence) on highly consistent target instances, and maximizes predictive entropy (reducing model confidence) on highly inconsistent ones.
    }
    \label{fig:approach}
    \vspace*{-10pt}
 \end{figure*}

In the second stage, the trained source model is adapted to the target with the use of unlabeled target and labeled source data. 
Recently, self-training via conditional entropy minimization (CEM)~\cite{grandvalet2005semi} has been shown to lead to strong performance for domain adaptation~\cite{saito2019semi}. This approach optimizes model parameters to minimize conditional entropy on unlabeled target data $\mathcal{H}_{\modelparams}(y|\mathbf{x})$. The entropy minimization objective $\mathcal{L}_{ENT}$ is given by:
\begin{equation}
    \vspace{-2pt}
\begin{aligned}        
    &  \mathcal{L}_{ENT} = \mathbb{E}_{\mathbf{x}_\target \sim \mathcal{P}_\target} \left[\mathcal{H}_{\modelparams}(y|\mathbf{x}_\target)\right] \\
        & = \mathbb{E}_{\mathbf{x}_\target \sim \mathcal{P}_\target} \left[\sum_{c=1}^{\numclasses} - p_{\modelparams}(y\!=\!c|\mathbf{x}_\target) \log p_{\modelparams}(y\!=\!c|\mathbf{x}_\target)\right]
\end{aligned}
\label{eq:ent}
\end{equation}
However, in many real-world scenarios, in addition to covariate shift, \emph{label distributions} across domains might also shift. Further, there might also be significant label imbalance within the target domain. In such cases, naive CEM has been found to potentially encourage trivial solutions of only predicting the majority class~\cite{li2020rethinking}. Li~\etal~\cite{li2020rethinking} regularize CEM with an ``information-entropy'' objective $\mathcal{L}_{IE}$ that encourages the model to make diverse predictions over unlabeled target instances. This is achieved by computing a distribution over classes predicted by the model for the last-$Q$ instances, denoted by $q(\hat{y})$, and updating parameters to maximize entropy over these predictions. This method is shown to help with domain alignment in the presence of label-distribution shift~\cite{li2020rethinking}~\footnote{The objective is referred to as ``mutual information maximization'' in Li \etal~\cite{li2020rethinking}}. $\mathcal{L}_{IE}$ is defined as:

\vspace{-5pt}
\begin{equation}
\begin{aligned}        
    \mathcal{L}_{IE} & = \mathbb{E}_{\mathbf{x}_\target \sim \mathcal{P}_\target} \left[\sum_{c=1}^{\numclasses} p_{\modelparams}(y\!=\!c|\mathbf{x}_\target) \log q(\hat{y}\!=\!c)\right]
\end{aligned}
\label{eq:ie}
\end{equation}

\noindent \textbf{CEM and error accumulation.} While conditional entropy minimization has been a part of many successful approaches for semi-supervised learning~\cite{grandvalet2005semi,berthelot2019mixmatch}, few-shot learning~\cite{dhillon2019baseline}, and more recently, UDA~\cite{saito2019semi,li2020rethinking}, it suffers from a key challenge in the case of domain adaptation. Intuitively, conditional entropy minimization encourages the model to make confident predictions on unlabeled target data. This makes its success highly dependent on its initialization. Under a good initialization, categories may be reasonably aligned across source and target domains after source training, and such self-training works well. However, under strong domain shifts, several categories may initially be misaligned across domains, often systematically so, and entropy minimization will only lead to \emph{reinforcing} such errors.

\vspace{-4pt}
\subsection{\method: Selective Entropy Optimization via Committee Consistency}
\vspace{-4pt}

\noindent \textbf{Predictive consistency-based selection.} To address the problem of error accumulation under CEM, we propose \emph{selective} optimization on well-aligned instances. The question then becomes: how can we identify reliable instances? One possibility is to use top-1 softmax confidence (or alternatively, predictive entropy), and only self-train on highly confident instances, as done in prior work~\cite{tan2019generalized}. However, under a distribution shift, such confidence measures tend to be miscalibrated and are often unreliable~\cite{snoek2019can}. Instead, we propose using \emph{predictive consistency} under a committee of label-preserving image transformations as a more robust measure for instance selection.

For a target instance $\mathbf{x}_\target \sim \mathcal{P}_{\target}$, we generate a committee of $k$ transformed versions $\{a_1(\mathbf{x}_\target), a_2(\mathbf{x}_\target), ..., a_k(\mathbf{x}_\target)\}$. 
We make predictions for each of these $k$ transformed versions, and measure \emph{consistency} between the model's prediction for the original image and for each of its $k$ augmented versions. In practice, we use a simple majority voting scheme: if the model's prediction for a majority of augmented versions matches its prediction on the original image, we consider the instance as ``consistent''. Similarly, if the prediction for a majority of augmented versions does not match the original prediction, we mark it as ``inconsistent''.

\noindent \textbf{Selective Entropy Optimization.} Having identified consistent and inconsistent instances, we perform Selective Entropy Optimization (\method). 
First, for an instance marked as consistent, we \emph{increase model confidence} by minimizing predictive entropy~\cite{grandvalet2005semi} with respect to one of its consistent augmented versions.

As described previously, some target instances may be misaligned under a domain shift. Entropy minimization on such instances would increase model confidence, \emph{reinforcing such errors}. Instead, having identified such an instance via predictive inconsistency, we \emph{reduce model confidence} by \emph{maximizing} predictive entropy~\cite{pereyra2017regularizing} with respect to one of its inconsistent augmented versions.
While the former encourages confident predictions on highly consistent instances, the latter reduces model confidence on highly inconsistent and likely misaligned instances. In Sec.~\ref{ref:analysis}, we provide further intuition into the behavior of entropy maximization by illustrating its similarity to a binary cross-entropy loss with respect to the ground truth label for an incorrectly classified example in the binary classification case.

Without loss of generality, we minimize/maximize entropy with respect to the last consistent/inconsistent transformed version in our experiments. Our selective entropy optimization objective $\mathcal{L}_{\method}$ is given by:

\vspace{-15pt}
\begin{equation}
\mathcal{L}_{\method}(\mathbf{x}_\target)= 
\begin{cases}
  + \mathcal{H}_{\modelparams}(y|a_i(\mathbf{x}_\target))  & \text{if consistent} \\
  - \mathcal{H}_{\modelparams}(y|a_j(\mathbf{x}_\target))  & \text{if inconsistent}
\vspace{-5pt}
\end{cases}
\end{equation}

\noindent Here $i$ and $j$ denote the index of the last consistent and inconsistent transformed version, respectively.

Such an approach may raise two concerns: First, that entropy minimization only on consistent instances might lead to the exclusion of a large percentage of target instances. Second, that indefinite entropy maximization on inconsistent target instances might prove detrimental to learning. Both of these concerns are addressed via the augmentation invariance regularizer built into our objective, which leads to an adaptive selection strategy that we now discuss.

\noindent \textbf{Adaptive selection via augmentation invariance regularization.} For instances marked as consistent, our approach minimizes entropy with respect to its last consistent \emph{augmented} version rather than with respect to the original image itself. This yields two benefits: First, this builds data augmentation into the entropy minimization objective, which helps reduce overfitting. Second, it encourages invariance to the same set of augmentations that is used for selecting instances. We find that this makes our selection strategy \emph{adaptive}, wherein an increasing percentage of target instances are selected for entropy minimization over the course of training, and consequently a decreasing percentage of target instances are selected for entropy maximization.

\begin{algorithm}
\caption{\method Optimization}
\label{algo:sentry}
\begin{algorithmic}[1]
\State Input: $\mathcal{X_\source}$, $\mathcal{Y_\source}$, $\mathcal{X_\target}$, $Q$, $\Theta$
\ForAll{$x_{T}^{(i)} \in \mathcal{X_\target}$} \Comment{\textcolor{blue}{Init target pseudo-labels}}
    \State $\hat{\mathcal{Y}}_\target^{(i)} \gets \argmax p_{\Theta}(y | x_\target^{(i)})$    
\EndFor
\State SrcLoader $\gets$ ClassBalancedSampler($\mathcal{X}_\source, \mathcal{Y_\source}$)
\State TgtLoader $\gets$ ClassBalancedSampler($\mathcal{X}_\target, \mathcal{\hat{Y}_\target}$)
\State $q \gets $ Queue(\texttt{size}=Q)
\For{epoch $\gets$ 1 to \texttt{MAX\_EPOCH}}
    \For{$x_\source,y_\source$ in SrcLoader and $x_\target$ in TgtLoader}
        \State $\hat{y}_\target \gets \argmax p_{\Theta}(y | x_\target)$\Comment{\textcolor{blue}{Clean prediction}}
        \State $\{a_1(x_\target), \dots, a_k(x_\target) \} \gets$ RandAugment($x_\target$)
        \State {\small $\text{C} \gets \{a_i(x_\target)| \hat{y}_\target = \argmax p_{\Theta}(y | a_i(x_\target))\}_{i=1}^k$}
        \State {\small  $\text{IC} \gets \{a_i(x_\target)| \hat{y}_\target \neq \argmax p_{\Theta}(y | a_i(x_\target))\}_{i=1}^k$}
        
        \If{\texttt{len}(\text{C}) $>$ \texttt{len}(\text{IC})} \Comment{\textcolor{blue}{Consistent}}
            \State $\mathcal{L}_{\texttt{SENTRY}} = \mathcal{H}_{\modelparams}(y | \text{C}\text{.last()})$

        \Else \Comment{\textcolor{blue}{Inconsistent}}
            \State $\mathcal{L}_{\texttt{SENTRY}} = -\mathcal{H}_{\modelparams}(y | \text{IC}\text{.last()})$
        \EndIf
        \State Update($\hat{Y}_\target, \hat{y}_\target)$
        \State $q$.enqueue($\hat{y}_\target$) \Comment{\textcolor{blue}{Update pseudo-label queue}}
    \EndFor
    \State Minimize $\mathcal{L}_{\texttt{SENTRY}} + \mathcal{L}_{IE}(q) + \mathcal{L}_{CE}(x_\source,y_\source) $
    \State TgtLoader $\gets$ ClassBalancedSampler($\mathcal{X}_\target, \mathcal{\hat{Y}_\target}$)
\EndFor
\end{algorithmic}
\end{algorithm}
\vspace*{-1.1cm}
\subsection{Overcoming LDS via pseudo class balancing} 
\label{sec:pcb}

Under LDS, methods often have to adapt in the presence of severe label imbalance. While label imbalance on the source domain often leads to poor performance on tail classes~\cite{cui2019class,wang2017learning}, adapting to an imbalanced target often results in poor performance on head classes~\cite{li2020rethinking,wu2019domain}. To overcome this, we employ a simple class-balanced sampling strategy. On the source domain, we perform class-balanced sampling using ground truth labels. On the target domain, we approximate the label distribution via \emph{pseudolabels}, and perform approximate class-balanced sampling~\cite{zou2018unsupervised}. 

Such balancing also complements the target information-entropy loss $L_{IE}$~\cite{li2020rethinking} (Eq.~\ref{eq:ie}). To recap, $L_{IE}$ encourages a uniform distribution over predictions. Under severe label imbalance, it is possible to sample highly label-imbalanced batches (with most instances belonging to head classes) and so encouraging a uniform distribution over predictions can adversely affect learning. However, our class-balanced sampling strategy reduces the probability of such a scenario, and we find that it consistently improves performance.

Algorithm~\ref{algo:sentry} details our full approach. The complete objective we optimize is given by:

\vspace{-10pt}
\begin{equation}
    \vspace{-10pt}
    \begin{aligned}    
    \argmin_{\modelparams} \quad & \mathbb{E}_{(\mathbf{x}_\source, y_\source) \stackrel{\text{bal}}{\sim} \mathcal{P}_\source} \mathcal{L}_{CE} \quad+ \\
    & \mathbb{E}_{\mathbf{x}_\target \stackrel{\text{pbal}}{\sim} \mathcal{P}_\target} \lambda_{IE} \mathcal{L}_{IE} + \lambda_{\method} \mathcal{L}_{\method}
    \end{aligned}    
    \end{equation}

\noindent where the $\lambda$'s denote loss weights, and $\stackrel{\text{bal}}{\sim}$ and $\stackrel{\text{pbal}}{\sim}$ denote balanced and pseudo class-balanced sampling.
\begin{figure*}[t]
    \centering
    \vspace{-3pt}
    \includegraphics[width=\textwidth]{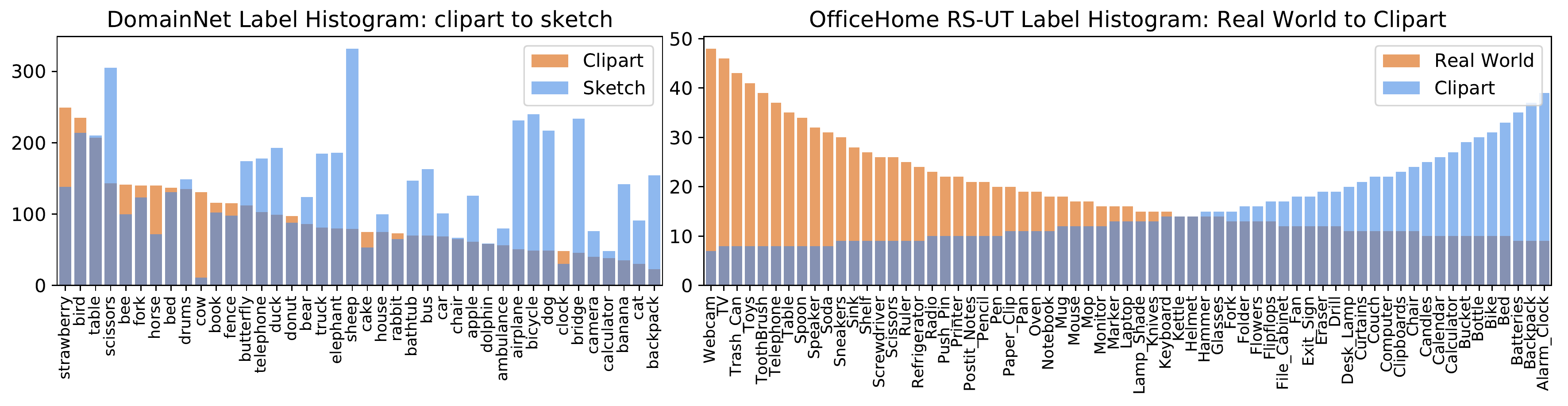}
    \vspace*{-17pt}
    \caption{
    \textbf{Left}: Natural label distribution shift (LDS) on the Clipart$\rightarrow$Sketch shift from DomainNet. \textbf{Right:} Manually generated LDS on the Real World$\rightarrow$Clipart shift from OfficeHome RS-UT~\cite{tan2019generalized}.
    }
    \vspace*{-7pt}
    \label{fig:dn_oh}
 \end{figure*}

 \vspace{-10pt}
\section{Experiments}
\label{sec:experiments}
\vspace{-5pt}

\par\noindent We first describe our experimental setup: datasets and metrics (Sec.~\ref{ref:datasets}), implementation details (Sec.~\ref{ref:implementation}), and baselines (Sec.~\ref{ref:baselines}). We then present our results (Sec.~\ref{ref:results}), ablation studies (Sec.~\ref{ref:ablations}), and analyze our approach (Sec.~\ref{ref:analysis}).

\vspace{-5pt}
\subsection{Datasets and Metrics}
\label{ref:datasets}
\vspace{-5pt}

\par\noindent We report results on a mix of standard UDA benchmarks and specialized benchmarks designed to stress-test UDA methods under label distribution shift.

\par\noindent\textbf{DomainNet.} DomainNet~\cite{peng2019moment} is a large UDA benchmark for image classification, containing 0.6 million images belonging to 6 domains spanning 345 categories. Due to labeling noise prevalent in its full version, we instead use the subset proposed in Tan~\etal~\cite{tan2019generalized}, which uses 40-commonly seen classes from 4 domains: Real (\textbf{R}), Clipart (\textbf{C}), Painting (\textbf{P}), and Sketch (\textbf{S}). As seen in Fig.~\ref{fig:dn_oh} (left), there exists a natural label distribution shift across domains, which makes it suitable for testing our method without manual subsampling.

\par\noindent\textbf{OfficeHome.} OfficeHome~\cite{venkateswara2017deep} is an image classification-based benchmark containing 65 categories of objects found in office and home environments, spanning 4 domains: Real-world (\textbf{Rw}), Clipart (\textbf{Cl}), Product (\textbf{Pr}), and Art (\textbf{Ar}). We report performance on two versions: i) standard: the original dataset proposed in Venkateswara~\etal~\cite{venkateswara2017deep}, and ii) RS-UT: The Reverse-unbalanced Source (RS) and Unbalanced-Target (UT) version from Tan~\etal~\cite{tan2019generalized}, wherein source and target label distributions are manually long-tailed to be reversed versions of one another (see Fig.~\ref{fig:dn_oh} (right)).

\par\noindent\textbf{VisDA.} VisDA2017~\cite{peng2017visda} is a large dataset for synthetic$\rightarrow$real adaptation with 12 classes and >200k images.

\begin{table*}[t]
    \begin{center} 
    \setlength{\tabcolsep}{2pt}
    \resizebox{\textwidth}{!}{
    \begin{tabular}{l c c c c c c c c c c c c c}
        \toprule
        Method & $\mathbf{R} \rightarrow \mathbf{C}$ & $\mathbf{R} \rightarrow \mathbf{P}$ & $\mathbf{R} \rightarrow \mathbf{S}$ & $\mathbf{C} \rightarrow \mathbf{R}$ & $\mathbf{C} \rightarrow \mathbf{P}$ & $\mathbf{C} \rightarrow \mathbf{S}$ & $\mathbf{P} \rightarrow \mathbf{R}$ & $\mathbf{P} \rightarrow \mathbf{C}$ & $\mathbf{P} \rightarrow \mathbf{S}$ & $\mathbf{S} \rightarrow \mathbf{R}$ & $\mathbf{S} \rightarrow \mathbf{C}$ & $\mathbf{S} \rightarrow \mathbf{P}$ & AVG \\
        \midrule
        source & 65.75 & 68.84 & 59.15 & 77.71 & 60.60 & 57.87 & 84.45 & 62.35 & 65.07 & 77.10 & 63.00 & 59.72 & 66.80 \\
        \midrule
        BBSE~\cite{lipton2018detecting} & 55.38 & 63.62 & 47.44 & 64.58 & 42.18 & 42.36 & 81.55 & 49.04 & 54.10 & 68.54 & 48.19 & 46.07 & 55.25 \\
        PADA~\cite{cao2018partial} & 65.91 & 67.13 & 58.43 & 74.69 & 53.09 & 52.86 & 79.84 & 59.33 & 57.87 & 76.52 & 66.97 & 61.08 & 64.48 \\
        MCD~\cite{saito2018maximum} & 61.97 & 69.33 & 56.26 & 79.78 & 56.61 & 53.66 & 83.38 & 58.31 & 60.98 & 81.74 & 56.27 & 66.78 & 65.42 \\
        DAN~\cite{long2015learning} & 64.36 & 70.65 & 58.44 & 79.44 & 56.78 & 60.05 & 84.56 & 61.62 & 62.21 & 79.69 & 65.01 & 62.04 & 67.07 \\
        F-DANN~\cite{wu2019domain} & 66.15 & 71.80 & 61.53 & 81.85 & 60.06 & 61.22 & 84.46 & 66.81 & 62.84 & 81.38 & 69.62 & 66.50 & 69.52 \\
        UAN~\cite{you2019universal} & 71.10 & 68.90 & 67.10 & 83.15 & 63.30 & 64.66 & 83.95 & 65.35 & 67.06 & 82.22 & 70.64 & 68.09 & 72.05 \\
        JAN~\cite{long2017deep} & 65.57 & 73.58 & 67.61 & 85.02 & 64.96 & 67.17 & 87.06 & 67.92 & 66.10 & 84.54 & 72.77 & 67.51 & 72.48 \\
        ETN~\cite{cao2019learning} & 69.22 & 72.14 & 63.63 & 86.54 & 65.33 & 63.34 & 85.04 & 65.69 & 68.78 & 84.93 & 72.17 & 68.99 & 73.99 \\
        BSP~\cite{chen2019transferability} & 67.29 & 73.47 & 69.31 & 86.50 & 67.52 & 70.90 & 86.83 & 70.33 & 68.75 & 84.34 & 72.40 & 71.47 & 74.09 \\
        DANN~\cite{ganin2014unsupervised} & 63.37 & 73.56 & 72.63 & 86.47 & 65.73 & 70.58 & 86.94 & 73.19 & 70.15 & 85.73 & 75.16 & 70.04 & 74.46 \\
        COAL~\cite{tan2019generalized} & 73.85 & 75.37 & 70.50 & 89.63 & 69.98 & 71.29 & \underline{89.81} & 68.01 & 70.49 & \underline{87.97} & 73.21 & 70.53 & 75.89 \\
        InstaPBM~\cite{li2020rethinking} & \underline{80.10} & \underline{75.87} & \underline{70.84} & \underline{89.67} & \underline{70.21} & \underline{72.76} & 89.60 & \underline{74.41} & \underline{72.19} & 87.00 & \underline{79.66} & \underline{71.75} & \underline{77.84} \\
        \midrule
        \method (Ours) & $\mathbf{83.89}$ & $\mathbf{76.72}$ & $\mathbf{74.43}$ & $\mathbf{90.61}$ & $\mathbf{76.02}$ & $\mathbf{79.47}$ & $\mathbf{90.27}$ & $\mathbf{82.91}$ & $\mathbf{75.60}$ & $\mathbf{90.41}$ & $\mathbf{82.40}$ & $\mathbf{73.98}$ & $\mathbf{81.39}$ \\        
        \bottomrule
        \end{tabular}}
        \vspace{-5pt}
        \caption{Per-class average accuracies on DomainNet. Bold and underscore denote the best and second-best performing methods respectively.}\label{tab:domainnet}
        \vspace{-10pt}
    \end{center}
    \end{table*}

    \begin{table}[t]
        \begin{center} 
        \setlength{\tabcolsep}{1.5pt}
        \resizebox{\columnwidth}{!}{
    \begin{tabular}{lccccccc}
        \toprule        
        Method & \small{$\mathbf{Rw}\veryshortarrow\mathbf{Pr}$} & \small{$\mathbf{Rw}\veryshortarrow\mathbf{Cl}$} & \small{$\mathbf{Pr}\veryshortarrow\mathbf{R w}$} & \small{$\mathbf{Pr}\veryshortarrow\mathbf{Cl}$} & \small{$\mathbf{Cl}\veryshortarrow\mathbf{R} \mathbf{w}$} & \small{$\mathbf{Cl}\veryshortarrow\mathbf{Pr}$} & AVG \\
        \midrule
        source & 70.74 & 44.24 & 67.33 & 38.68 & 53.51 & 51.85 & 54.39 \\
        \midrule
        BSP~\cite{chen2019transferability} & 72.80 & 23.82 & 66.19 & 20.05 & 32.59 & 30.36 & 40.97 \\
        PADA~\cite{cao2018partial} & 60.77 & 32.28 & 57.09 & 26.76 & 40.71 & 38.34 & 42.66 \\
        BBSE~\cite{lipton2018detecting} & 61.10 & 33.27 & 62.66 & 31.15 & 39.70 & 38.08 & 44.33 \\
        MCD~\cite{saito2018maximum} & 66.03 & 33.17 & 62.95 & 29.99 & 44.47 & 39.01 & 45.94 \\
        DAN~\cite{long2015learning} & 69.35 & 40.84 & 66.93 & 34.66 & 53.55 & 52.09 & 52.90 \\
        F-DANN~\cite{wu2019domain} & 68.56 & 40.57 & 67.32 & 37.33 & 55.84 & 53.67 & 53.88 \\
        JAN~\cite{long2017deep}  & 67.20 & 43.60 & 68.87 & 39.21 & 57.98 & 48.57 & 54.24 \\
        DANN~\cite{ganin2014unsupervised} & 71.62 & 46.51 & 68.40 & 38.07 & 58.83 & 58.05 & 56.91 \\
        MDD~\cite{zhang2019bridging} & 71.21 & 44.78 & 69.31 & 42.56 & 52.10 & 52.70 & 55.44 \\
        COAL~\cite{tan2019generalized} & 73.65 & 42.58 & 73.26 & 40.61 & 59.22 & 57.33 & 58.40 \\
        InstaPBM~\cite{li2020rethinking} & 75.56 & 42.93 & 70.30 & 39.32 & \underline{61.87} & \underline{63.40} & 58.90 \\
        MDD+I.A~\cite{jiang2020implicit} & \underline{76.08} & \underline{50.04} & $\mathbf{74.21}$ & \underline{45.38} & 61.15 & 63.15 & \underline{61.67} \\
        \midrule
        \method (Ours) & $\mathbf{76.12}$ & $\mathbf{56.80}$ & \underline{73.60} & $\mathbf{54.75}$ & $\mathbf{65.94}$ & $\mathbf{64.29}$ & $\mathbf{65.25}$ \\
        \bottomrule
        \end{tabular}
        }
        \vspace{-5pt}
        \caption{Per-class average accuracies on OfficeHome RS$\rightarrow$UT (right) benchmarks. Bold and underscore denote the best and second-best performing methods respectively.}
        \vspace{-10pt}
        \label{tab:officehome}
        \end{center}
    \end{table}

\par\noindent\textbf{DIGITS.} We use the SVHN~\cite{netzer2011reading}$\rightarrow$MNIST~\cite{lecun1998gradient} shift for 10-way digit recognition.

\par\noindent\textbf{Metric.} On LDS DA benchmarks (DomainNet and OfficeHome RS-UT), consistent with prior work in UDA under LDS~\cite{tan2019generalized,jiang2020implicit}, we compute a mean of per-class accuracy on the target test split as our metric, that weights performance on all classes equally. On standard DA benchmarks (OfficeHome and VisDA2017) we report standard accuracy. 

\vspace{-5pt}
\subsection{Implementation details}
\label{ref:implementation}
\vspace{-5pt}

\noindent We use PyTorch~\cite{paszke2019pytorch} for all experiments. On DomainNet, OfficeHome, and VisDA2017, we modify the standard ResNet50~\cite{he2016deep} CNN architecture to a few-shot variant used in recent DA work~\cite{chen2018closer,saito2019semi,tan2019generalized}: we replace the last linear layer with a $C-$ way (for $C$ classes) fully-connected layer with Xavier-initialized weights and no bias. We then $L_2$-normalize activations flowing into this layer and feed its output to a softmax layer with a temperature $T=0.05$. We match optimization details to Tan~\etal~\cite{tan2019generalized}. On DIGITS, we make similar modifications to the LeNet architecture and use $T=0.01$~\cite{hoffman2017cycada}. For augmenting images for consistency checking, we use RandAugment~\cite{cubuk2020randaugment}, which sequentially applies $N$ label-preserving image transformations randomly sampled from a set of 14 transforms. We set $N=3$, use transformation severity $M=2.0$, and use a committee of $k=3$ transforms. We use class-balanced sampling on the source domain and pseudo class-balanced sampling on the target. We set $\lambda_{IE}$ and $\lambda_{\method}$ to 0.1 and 1.0, and match InstaPBM to set Q$=$256 for the information entropy loss.

\vspace{-5pt}
\subsection{Baselines}
\label{ref:baselines}
\vspace{-5pt}

\noindent As our primary baselines we use four state-of-the art UDA methods from prior work specifically designed for DA under LDS: i) \textbf{COAL}~\cite{tan2019generalized}: Co-aligns feature and label distributions, using prototype-based conditional alignment via MME~\cite{saito2019semi}, and self-training on confidently-predicted pseudo-labels. ii) \textbf{MDD + Implicit Alignment (I.A)}~\cite{jiang2020implicit}: Uses target pseudolabels to construct $N-$way (\# classes per-batch) $K-$shot (\# examples per class) dataloaders that are ``aligned'' (i.e. sample the same set of classes within a batch for both source and target), in conjunction with Margin Disparity Discrepancy~\cite{zhang2019bridging}, a strong UDA method, iii) \textbf{InstaPBM~\cite{li2020rethinking}}: Proposes ``predictive-behavior'' matching, which entails matching properties of $p_{\modelparams}(y|\mathbf{x})$ between source and target. This is achieved via optimizing a combination of mutual information maximization, contrastive, and mixup losses, and iv) \textbf{F-DANN}~\cite{wu2019domain}: Proposes an asymmetrically-relaxed distribution matching-based version of DANN~\cite{ganin2014unsupervised} to deal with LDS. COAL, InstaPBM, and MDD+I.A. all make use of target pseudolabels, and COAL and InstaPBM are self-training based approaches.

For completeness, we also include results for additional baselines from Tan~\etal~\cite{tan2019generalized}: i) Conventional feature alignment-based UDA methods: DAN~\cite{long2015learning}, JAN~\cite{long2017deep}, DANN~\cite{ganin2014unsupervised}, MCD~\cite{saito2019semi}, and MDD~\cite{zhang2019bridging}, ii) BBSE~\cite{li2020rethinking} which only aligns label distributions, iii) Methods that assume non-overlapping labeling spaces: PADA~\cite{cao2018partial}, ETN~\cite{cao2019learning}, and UAN~\cite{you2019universal}. We also report results for FixMatch~\cite{sohn2020fixmatch}, a state-of-the-art self-training method for semi-supervised learning, on two benchmarks.

\vspace{-5pt}
\subsection{Results}
\label{ref:results}

\noindent \textbf{Results on label-shifted DA benchmarks.} We present results on 12 shifts from DomainNet (Table~\ref{tab:domainnet}) and 6 shifts from OfficeHome RS$\rightarrow$UT (Table~\ref{tab:officehome}). On DomainNet, \method outperforms the next best performing method InstaPBM~\cite{li2020rethinking} on every shift, and by 3.55\% mean accuracy averaged across shifts. On OfficeHome RS-UT, \method outperforms the next best performing method MDD+I.A~\cite{jiang2020implicit} on 5 out of 6 shifts, and on average by 3.58\% mean accuracy. Our method also significantly outperforms F-DANN~\cite{wu2019domain} (by 11.87\% and 11.37\%) and COAL~\cite{tan2019generalized} (by 5.50\% and 6.85\%), which are both UDA strategies designed for adaptation under LDS.

\begin{table}[h]    
    \centering
    \RawFloats
    \begin{subfloatrow} 
    \ffigbox[\FBwidth][][!htbp]    
    {
        \setlength{\tabcolsep}{3pt}
        \begin{tabular}{lc}
            \toprule
            Method & acc (\%) \\
            \midrule
            Source & 46.1 \\
            \midrule
            DAN~\cite{long2015learning} & 56.3 \\
            DANN~\cite{ganin2014unsupervised} & 57.6 \\
            JAN~\cite{long2017deep} & 58.3 \\
            CDAN~\cite{long2018conditional} & 65.8 \\
            BSP~\cite{chen2019transferability} & 66.3\\
            MDD~\cite{zhang2019bridging} & 68.1 \\
            FixMatch~\cite{sohn2020fixmatch} & 59.0 \\
            InstaPBM~\cite{li2020rethinking} & 69.2 \\              
            MDD+I.A~\cite{jiang2020implicit} & 69.5 \\
            \midrule
            \method (Ours) & \textbf{72.2} \\
            \bottomrule
            \end{tabular}
  }
  {
    \caption{\small OfficeHome (12 shift avg)}\label{tab:officehome_std}      
    }
\hspace{-1.0cm}
\ffigbox[\FBwidth][][b]
{
        \setlength{\tabcolsep}{3pt}
        \begin{tabular}{l c c c c}
            \toprule
            Method & acc (\%) \\
            \midrule
            Source & 41.0  \\    
            \midrule
            JAN~\cite{long2015learning} & 61.6  \\    
            MCD~\cite{saito2018maximum} & 69.8  \\    
            CDAN~\cite{long2018conditional} & 70.0  \\    
            FixMatch~\cite{sohn2020fixmatch} & 64.9 \\
            MDD~\cite{zhang2019bridging} & 74.6  \\    
            MDD+I.A~\cite{jiang2020implicit} & 75.8  \\    
            InstaPBM~\cite{li2020rethinking} & 76.3  \\    
            \midrule
            \method (Ours) & \textbf{76.7}  \\    
            \bottomrule
            \end{tabular}
  }
    {
    \caption{\small VisDA2017}
    \label{tab:visda}
    }
    \end{subfloatrow}
    \vspace{-5pt}
    \caption{Accuracies on standard DA benchmarks.}
    \vspace{-5pt}
    \label{tab:standard_da}
\end{table}

\noindent \textbf{Results on standard DA benchmarks.} Table~\ref{tab:standard_da} presents results on 2 standard DA benchmarks: OfficeHome and VisDA 2017. As seen, \method improves mean accuracy over the next best method by 2.7\% averaged over 12 shifts (full table in supp.) on OfficeHome, and by 0.4\% on VisDA.

\begin{table*}[t]    
    \centering
    \RawFloats
    \begin{subfloatrow} 
    \ffigbox[\FBwidth][][h]    
    {
        \setlength{\tabcolsep}{4pt}
        \begin{tabular}{l c c c c c}
            \toprule
            & \multicolumn{5}{c}{SVHN$\rightarrow$MNIST-LT} \\
            Method & \texttt{IF=1} & \texttt{IF=20} & \texttt{IF=50} & \texttt{IF=100} & AVG \\
            \midrule
            source & 68.1 & 68.1 & 68.1 & 68.1 & 68.1 \\
            \hline
        MMD~\cite{long2013transfer} & 53.4\scriptsize{$\pm$0.9} & 56.7\scriptsize{$\pm$1.2} & 56.2\scriptsize{$\pm$1.4} & 55.1\scriptsize{$\pm$0.7} & 55.4\scriptsize{$\pm$1.1}\\
            DANN~\cite{ganin2014unsupervised} & 68.0\scriptsize{$\pm$0.9} & 71.5\scriptsize{$\pm$1.0} & 66.9\scriptsize{$\pm$0.5} & 60.6\scriptsize{$\pm$2.2} & 66.8\scriptsize{$\pm$1.5} \\
            COAL~\cite{tan2019generalized} & 78.8\scriptsize{$\pm$1.0}& 67.1\scriptsize{$\pm$1.4} & 70.2\scriptsize{$\pm$1.5} & 70.0\scriptsize{$\pm$1.8} & 71.5\scriptsize{$\pm$1.4} \\
            InstaPBM~\cite{li2020rethinking} & 90.7\scriptsize{$\pm$0.2} & 77.9\scriptsize{$\pm$3.5} & 68.9\scriptsize{$\pm$1.3} & 65.9\scriptsize{$\pm$2.2} & 75.9\scriptsize{$\pm$1.8}\\
            \hline
            \method (Ours) & \textbf{92.9\scriptsize{$\pm$0.3}} & \textbf{93.9\scriptsize{$\pm$2.2}} & \textbf{85.6$\pm$\scriptsize{4.5}} & \textbf{85.6\scriptsize{$\pm$1.1}} & \textbf{89.5\scriptsize{$\pm$2.0}} \\
            \bottomrule
        \end{tabular}
  }
  {
    }
\ffigbox[\FBwidth][][h]
    {\includegraphics[width=0.35\textwidth]{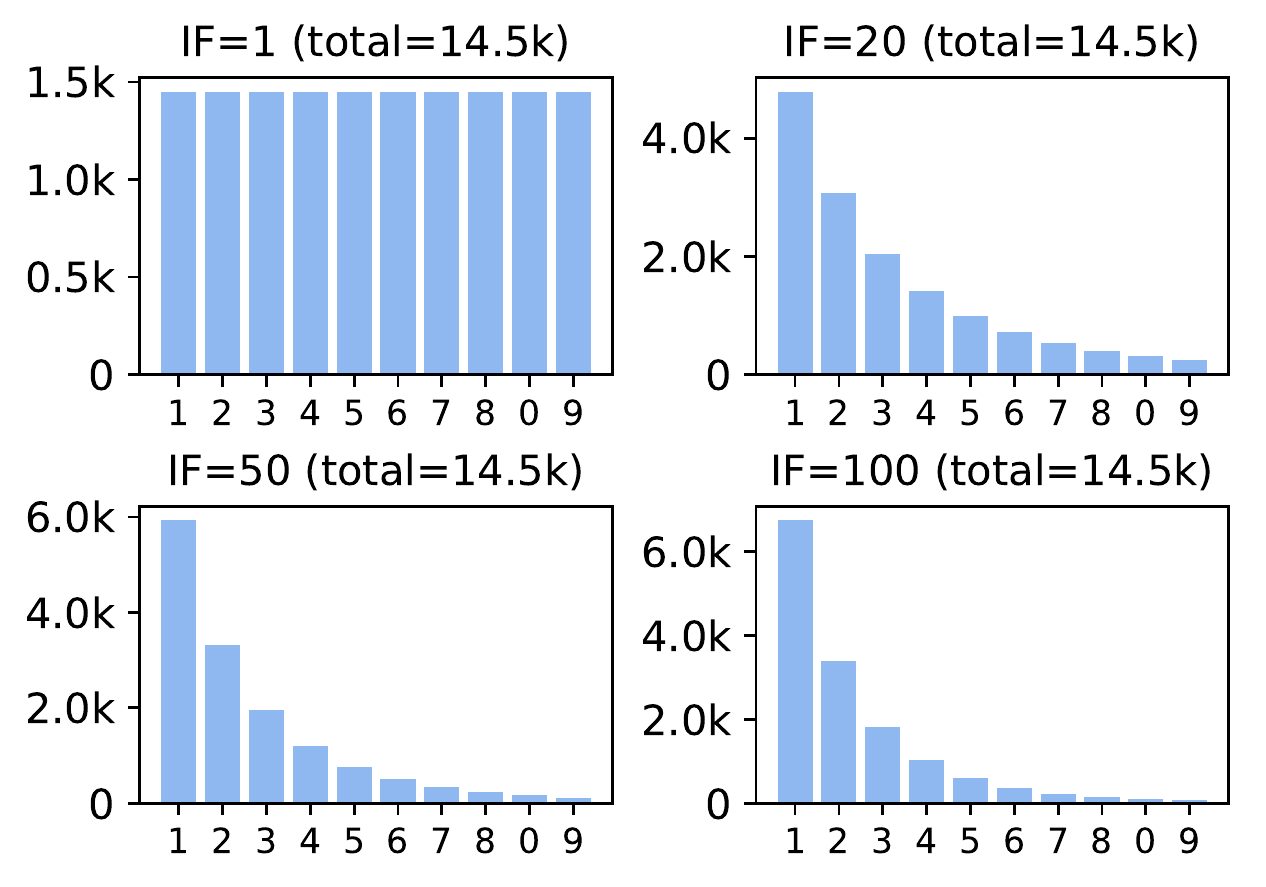}}
    {
    }
    \end{subfloatrow}
    \vspace{-7pt}
    {
    \caption{\textbf{Left}: Per-class average accuracy after adapting from SVHN to manually long-tailed (-LT) training sets of MNIST (test set is unchanged). The degree of label imbalance is measured by the imbalance factor (IF). All long-tailed versions use an \emph{identical} amount of data. For each IF, we construct 3 long-tailed versions and report mean and 1 standard deviation. \textbf{Right}: Label distribution for each IF.}
    \vspace{-5pt}
    \label{tab:digits}
}
\end{table*}

\noindent \textbf{Varying degree of label imbalance.} To perform a controlled study of adapting to targets with varying degrees of label imbalance, we use the SVHN$\rightarrow$MNIST shift. Since MNIST is class-balanced, we manually long-tail its training split, and use it as our unlabeled target train set (test set is unchanged). The long-tailing is performed by sampling from a Pareto distribution and subsampling, with class cardinality following the same sorted order as the source label distribution for simplicity. To systematically vary the degree of imbalance, we modulate the parameters of the Pareto distribution so as to generate a desired Imbalance Factor (IF)~\cite{cui2019class}, computed as the ratio of the cardinality of the largest and smallest classes. Larger IF's represent a higher degree of imbalance. We thus create 3 splits with IF $\in \{20, 50, 100\}$, corresponding to varying label imbalance but with an identical amount of data (=14.5k instances). Further, we create a \emph{control} version that also has 14.5k instances but possesses a balanced label distribution. Table~\ref{tab:digits} (right) shows the resulting label distributions.

We report per-class average accuracies in Table~\ref{tab:digits} (left). As baselines, we include a domain discrepancy based method (MMD~\cite{long2013transfer}), an adversarial DA method (DANN~\cite{ganin2014unsupervised}), as well as COAL~\cite{tan2019generalized} and InstaPBM~\cite{li2020rethinking}.
Across methods, performance at higher imbalance factors is worse, illustrating the difficulty of adapting under severe label imbalance. However, \method significantly outperforms baselines, even at higher imbalance factors, achieving 13.6\% higher mean accuracy than the next competing method. 

\vspace{-5pt}
\subsection{Ablations}
\label{ref:ablations}

\noindent We now present ablations of \method, our proposed approach, on the Clipart$\rightarrow$Sketch from DomainNet and the Real World$\rightarrow$Clipart shift from OfficeHome RS-UT.

\begin{table}[h]
    \begin{center} 
    \resizebox{\columnwidth}{!}{
    \begin{tabular}{l l c c}
    \toprule
    select for& select for& DomainNet & OH (RS-UT) \\
    entmin & entmax & C$\rightarrow$S & Rw$\rightarrow$Cl \\
    \midrule
   all & none & 68.8 & 44.9 \\
   consistent & none & 77.7 & 55.3 \\
   \rowcolor{Gray}
   consistent & inconsistent & 79.5 & 56.8 \\
   \midrule
   correct & none & 84.3 & 77.7 \\
   correct & incorrect & 86.3 & 80.1 \\
    \bottomrule
    \end{tabular}}
    \vspace{-10pt}
    \caption{\small Ablations of our selection strategy on DomainNet C$\rightarrow$S and OfficeHome RS-UT Rw$\rightarrow$Cl. Gray row is our method. Last two rows are oracle approaches that use target labels.}
    \label{tab:ablate_ent}
    \vspace*{-5pt}
    \end{center}
\end{table}

\begin{table}[h]    
    \centering
    \RawFloats
    \begin{subfloatrow} 
    \hspace{-0.1cm}
    \ffigbox[\FBwidth][][!htbp]    
    {
    \resizebox{.32\textwidth}{!}{
        \setlength{\tabcolsep}{3pt}
        \begin{tabular}{c | c c c}
            \toprule
            & $N$=1& $N$=3 & $N$=5\\
            \hline
           $k$=1 & 78.2 & 78.6 & 78.9 \\
           $k$=3 & 76.8 & \cellcolor{Gray}79.5 & 77.8 \\
           $k$=5 & 77.5 & 78.4 & 77.7 \\
            \bottomrule
            \end{tabular}        
  }
  }
  {
      \caption{\small C$\rightarrow$S}
      \label{tab:ablate_con_dn}
    }
\hspace{-1.3cm}
\ffigbox[\FBwidth][][!htbp]
{
    \resizebox{.32\textwidth}{!}{
        \setlength{\tabcolsep}{3pt}
        \begin{tabular}{c | c c c}
            \toprule
            & $N$=1& $N$=3 & $N$=5\\
            \hline
            $k$=1 & 53.8 & 57.5 & 55.6 \\
            $k$=3 & 55.3 & \cellcolor{Gray}56.8 & 56.2 \\
            $k$=5 & 54.7 & 58.4 & 54.5 \\
            \bottomrule
            \end{tabular}
  }}
    {
    \caption{\small Rw$\rightarrow$Cl}
    \label{tab:ablate_con_oh}
    }
\hspace{-0.4cm}
    \ffigbox[\FBwidth][][b]
{
    \resizebox{.32\textwidth}{!}{
        \setlength{\tabcolsep}{3pt}
        \begin{tabular}{c c c}
            \toprule
            voting  & C$\rightarrow$S & Rw$\rightarrow$Cl \\
            \midrule
            \rowcolor{Gray}
            maj. & 79.5 & 56.8 \\
            unan. & 77.8 & 52.2 \\
            \bottomrule
            \end{tabular}
  }}
    {
    \caption{\small Vary voting}
    \label{tab:ablate_con_vote}
    }
    \end{subfloatrow}
    \caption{\small Ablating the consistency checker on C$\rightarrow$S and Rw$\rightarrow$Cl: \textbf{a-b)} Varying committee size ($k$) and num. consecutive transforms in RandAugment ($N$). \textbf{c)} Varying voting strategy: maj. and unan. denote majority and unanimous. Gray is our method.}
    \label{tab:ablate_con}
\end{table}

\noindent \textbf{Selective optimization helps significantly (Tab.~\ref{tab:ablate_ent})}. We first measure the effect of performing entropy minimization on \emph{all samples}, as done in prior work. We find this to perform \emph{significantly} worse (by 10.7\%, 11.9\%) than our method! Clearly, consistency-based selective optimization is crucial. 

\noindent \textbf{Entropy maximization helps consistently (Tab.~\ref{tab:ablate_ent})}. Next, we opt to only minimize entropy on consistent target instances, but \emph{do not perform entropy maximization}. We find this to underperform against our min-max optimization (by 1.8\%, 1.5\%).
Further, as an oracle approach, we use ground truth target labels to determine whether an instance is correctly or incorrectly classified, and perform two experiments: entropy minimization on correct instances (and no maximization), and min-max entropy optimization on correct and incorrect instances. Selective min-max optimization again outperforms just minimization by 2\% and 2.4\%, showing that reducing confidence on misaligned instances helps.

\noindent \textbf{Ablating consistency checker}. In Tables~\ref{tab:ablate_con_dn},~\ref{tab:ablate_con_oh}, we vary the committee size $k$ and number of RandAugment transforms $N$ used by our consistency checker. We do not find our method to be very sensitive to either hyperparameter. In Table~\ref{tab:ablate_con_vote}, we vary the voting strategy used to judge committee consistency and inconsistency. We experiment with majority voting (atleast $\frac{k}{2}+1$ votes needed) and unanimous voting ($k$ votes needed), and find the former to generalize better.

\begin{figure*}[t]
    \centering
    \includegraphics[width=1.0\textwidth]{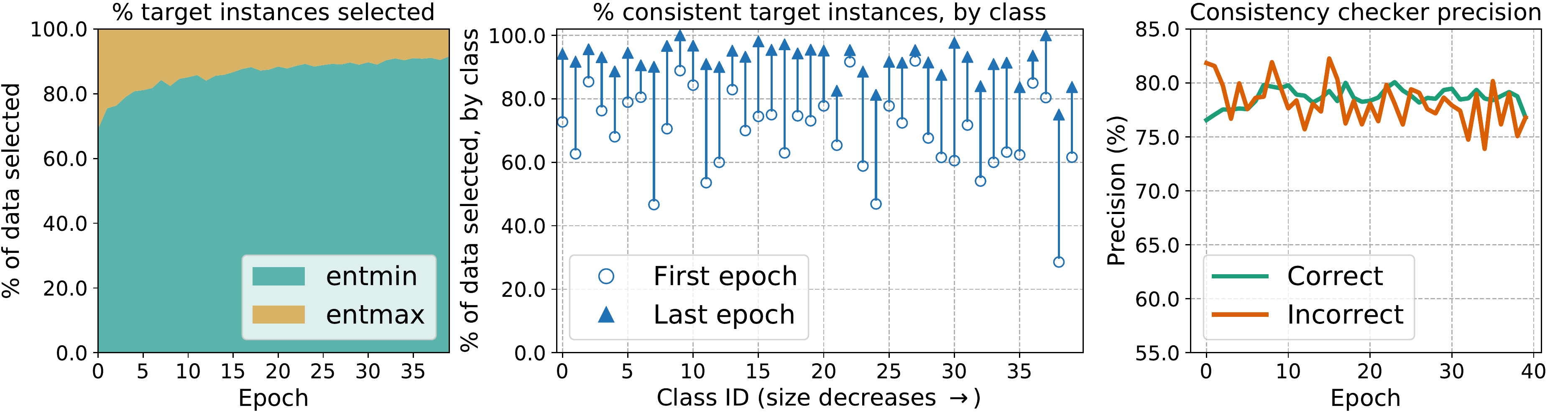}
    \vspace{-8pt}
    \caption{
    Analysis of \method on Clipart$\rightarrow$Sketch. \textbf{Left:} \% of seen target instances selected for entropy minimization and maximization over epochs. \textbf{Middle:} \% of seen target data chosen for entropy minimization at the end of first and last epochs of adaptation, broken down by class. \textbf{Right:} Ground truth precision of \method's committee consistency strategy at identifying correct and incorrect instances over epochs.
    }
    \vspace{-15pt}
    \label{fig:consistency_perf}
 \end{figure*}

 \noindent \textbf{Gains are not simply due to stronger augmentation.} To verify this, we continue using RandAugment (with N=1) for consistency checking, but backpropagate on the original (rather than augmented) target instances, effectively removing data augmentation entirely. On C$\to$S and Rw$\to$Cl, this achieves 73.1\% and 52.6\%, which is still 0.3\% and 2.6\% better than the next best baseline on each shift, despite not using any data augmentation for optimization at all.

\noindent \textbf{Pseudo class-balanced sampling helps.} We find that class-balanced sampling using pseudolabels on the target improves per-class average accuracy over random sampling by 0.91\% and 0.52\% on C$\rightarrow$S and Rw$\rightarrow$Cl. In the absence of the target information entropy regularizer $L_{IE}$, this performance gap grows to 2.9\% and 3.7\%.
As explained in Sec~\ref{sec:pcb}, both objectives contribute towards overcoming LDS in similar ways, and we find here that using both together works best.

\vspace{-5pt}
\subsection{Analysis}
\label{ref:analysis}
\vspace{-2pt}

\par\noindent \textbf{\% of target instances selected over time}. Fig.~\ref{fig:consistency_perf} (left) shows that the \% of seen target instances selected for entropy minimization steadily increases over time, while that selected for entropy maximization decreases. This adaptive nature is a result of the augmentation invariance regularization built into our method (Sec.~\ref{sec:approach}). In Fig.~\ref{fig:consistency_perf} (middle), we measure the \% of target instances selected for entropy minimization, \emph{per-class}, at the end of the first and last epoch of adaptation. Despite no explicit class-conditioning, we find that this measure increases for all classes.

\par\noindent \textbf{Precision of consistency checker.} Fig~\ref{fig:consistency_perf} (right) shows the precision of our consistency and inconsistency-based selection strategies at identifying instances that are actually correct and incorrect. As seen, committee-based consistency and inconsistency are both 75-80\% precise at identifying correct and incorrect instances respectively.

\par\noindent \textbf{Per-class accuracy change.} In the supplementary, we report the per-class accuracy after adaptation using our method, and contrast it against InstaPBM~\cite{li2020rethinking}. On the C$\to$S shift, \method outperforms InstaPBM across 37/40 categories.

\par\noindent \textbf{Computational efficiency.} Compared to standard entropy minimization, \method requires $k$ (for committee size $k$) forward passes per iteration (to determine consistency), but no additional backward passes. \method thus does not add a sizeable computational overhead over prior work.

\begin{wrapfigure}{r}{1.3in}
    \vspace{-\intextsep}
    \includegraphics[width=\linewidth]{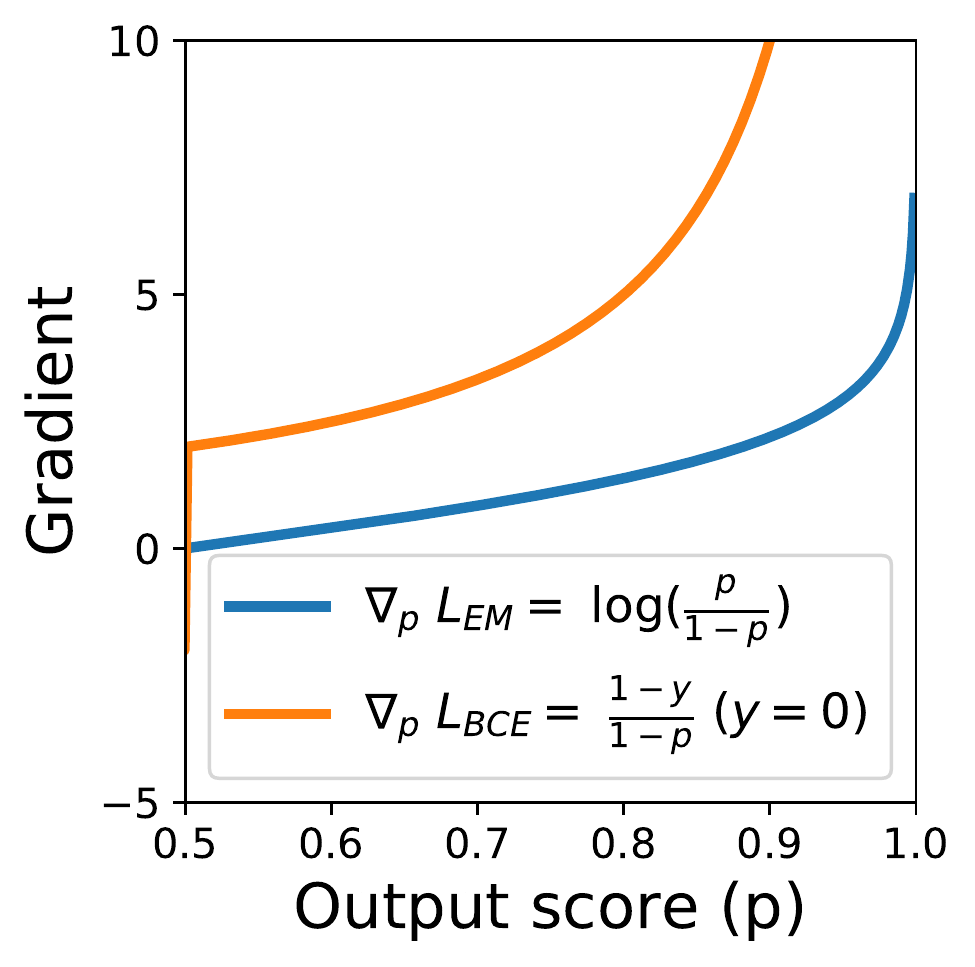}
    \caption{$\nabla_{p} L$ v/s p}%
        \vspace{-10pt}
    \label{fig:grad_corr}
  \end{wrapfigure}
\par\noindent \textbf{Correctness of entropy maximization.} 
For simplicity, consider 2-way classification. For output score $p$ and true class $y$, entropy maximization loss $L_{\text{EM}}=p\log p + (1-p)\log(1-p)$, and binary cross-entropy (BCE) loss $L_{BCE} = -\left[ y\log(p) + (1-y)\log(1-p)\right]$. Without loss of generality, assume an incorrect prediction with $y=0$ and $0.5 \le p < 1$. In Fig.~\ref{fig:grad_corr} we show that in this case, gradients ($\nabla_{p} L$) for entropy maximization and BCE (wrt true class) are strongly correlated. Thus, after identifying misaligned target instances based on predictive inconsistency, entropy maximization has a similar effect as supervised training with respective to the true class.

\vspace{-7pt}
\section{Conclusion}
\vspace{-5pt}

\noindent We propose \method, an algorithm for unsupervised domain adaptation (UDA) under simultaneous data and label distribution shift. Unlike prior work that suffers from error accumulation arising from unconstrained self-training, \method first judges the reliability of a target instance based on its predictive consistency under a committee of random image transforms, and then selectively minimizes entropy (increasing confidence) on consistent instances, while maximizing entropy (reducing confidence) on inconsistent ones. We show that \method significantly improves upon the state-of-the-art across 27/31 shifts from several UDA benchmarks.

\noindent\textbf{Acknowledgements.} This work was
supported in part by funding from the DARPA LwLL project.

{\small
\bibliographystyle{ieee_fullname}
\bibliography{main}
}

\section{Appendix}
\localtableofcontents

\newcommand\DoToC{%
  \startcontents
  \printcontents{}{2}{\textbf{Contents}\vskip3pt\hrule\vskip5pt}
  \vskip3pt\hrule\vskip5pt
}

\begin{figure*}[b]
    \centering
    \includegraphics[width=0.8\textwidth]{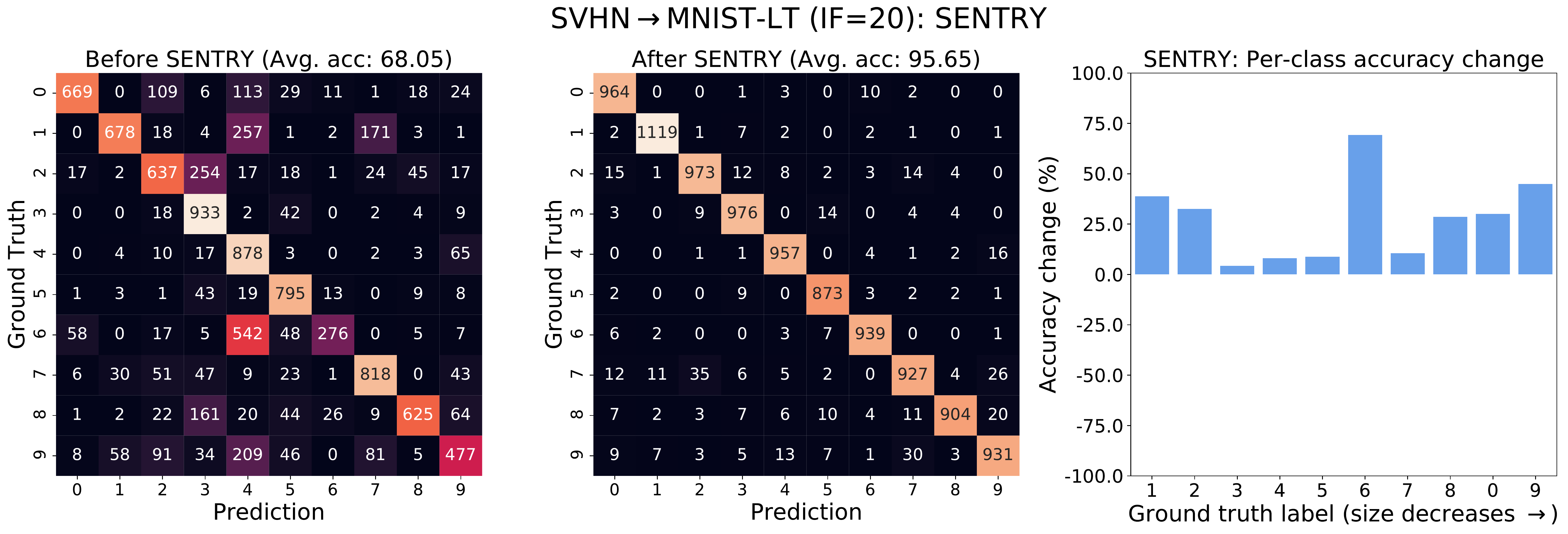}
    \caption{
    SVHN$\rightarrow$MNIST-LT (IF=20): Performance on target test set after \method.
    }
    \label{fig:sentry_cm}
 \end{figure*}

\subsection{Ablating class-balancing}

In Tab.~\ref{tab:ablate_cb}, we include results for ablating class-balancing on both the source and target domains. As seen, both contribute a small gain: +0.5\%, +0.3\% on C$\to$S (Row 1 v/s 2) for source class-balancing and +2\%, +0.1\% on Rw$\to$Cl (Row 1 v/s 3) for target pseudo class-balancing. Using both together works best (Row 4). However, even \emph{without any class balancing} (Row 1), \method is still 3.8\% and 6.2\% better than the next best method on each shift (Tab. 1-2 in paper). This confirms that the gains are due to \method's predictive consistency-based selective optimization.

\begin{table}[t]
    \begin{center} 
    \resizebox{0.99\columnwidth}{!}{
    \begin{tabular}{l c c c c }
    \toprule
    \#& CB & pseudo CB& DomainNet & OH (RS-UT) \\
    & (source) & (target) & {\centering C$\rightarrow$S} & Rw$\rightarrow$Cl \\
    \midrule
    1 &&  & 76.6 & 56.2 \\
    \midrule
    2&\ding{51} &  & 77.1$_{\textcolor{black}{+0.5}}$ & 56.5$_{\textcolor{black}{+0.3}}$ \\
    3&& \ding{51} & 78.6$_{\textcolor{black}{+2.0}}$ & 56.3$_{\textcolor{black}{+0.1}}$ \\
   \rowcolor{Gray}
   4&\ding{51} & \ding{51} & 79.5$_{\textcolor{black}{+2.9}}$ & 56.8$_{\textcolor{black}{+0.6}}$ \\
    \bottomrule
    \end{tabular}
    }
    \caption{Ablating class balancing. Gray row=\method. CB=class balancing. subscript=improvement v/s row 1.}
  
    \label{tab:ablate_cb}   
    \end{center}
    \vspace{-15pt}
 \end{table}

\subsection{Role of FS-architecture} 
 
\noindent Recall that in our experiments, we matched the ``few-shot'' style CNN architecture used in Tan~\etal~\cite{tan2019generalized} (Sec 4.2 in main paper). We now quantify the effect of this choice. As before, we measure average accuracy on DomainNet Clipart$\rightarrow$Sketch (C$\rightarrow$S) and OfficeHome RS-UT Real World$\rightarrow$Clipart (Rw$\rightarrow$Cl). We rerun our method without the few-shot modification, and observe a 1.6\% drop on C$\rightarrow$S and a 0.03\% increase in average accuracy on Rw$\rightarrow$Cl. Overall, this modification seems to lead to a slight gain.

\subsection{Additional analysis} 

\par
\begin{figure}[H]
    \centering
    \includegraphics[width=\columnwidth]{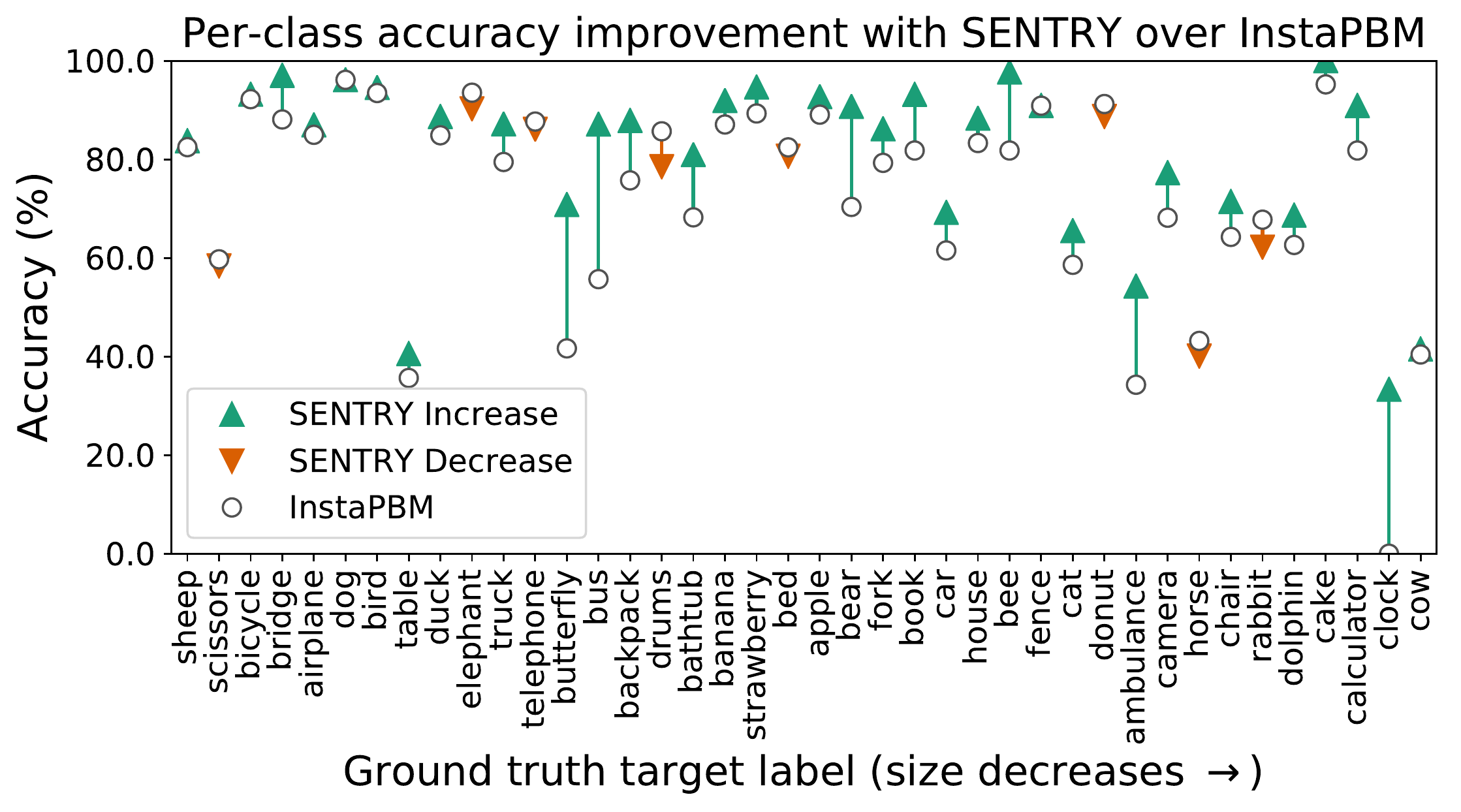}
    \caption{{\small
    DomainNet C$\rightarrow$S: Per-class accuracy gain with \method over InstaPBM. Classes are sorted by size (largest $\veryshortarrow{}$ smallest). }}
    \label{fig:pcadiff}
    \vspace{-10pt}
 \end{figure}

\noindent \textbf{Per-class accuracy change.} In Fig.~\ref{fig:pcadiff}, we report the per-class accuracy (sorted by class cardinality) after adaptation using our method on DomainNet Clipart$\to$Sketch, and contrast it against the next best-performing method, InstaPBM~\cite{li2020rethinking}. As seen, \method outperforms InstaPBM on 37/40 categories, and is competitive on the others.

We further analyze the performance of \method on the SVHN$\rightarrow$MNIST-LT (IF=20) shift, wherein the target train set has been manually long-tailed to create an imbalance factor of 20 (Sec 4.4 of main paper). In Fig~\ref{fig:sentry_cm}, we show a confusion matrix of model predictions on the target test set after source training (left) and after target adaptation via \method (middle). As seen, strong misalignments exist initially. However, after adaptation via our method, alignment improves dramatically across all classes. In Fig~\ref{fig:sentry_cm} (right), we show the \emph{change} in per-class accuracy after adaptation, while sorting classes in decreasing order of size. As seen, \method improves performance for both head and tail classes, often very significantly so.

\begin{figure*}[t]
    \centering
    \includegraphics[width=\textwidth]{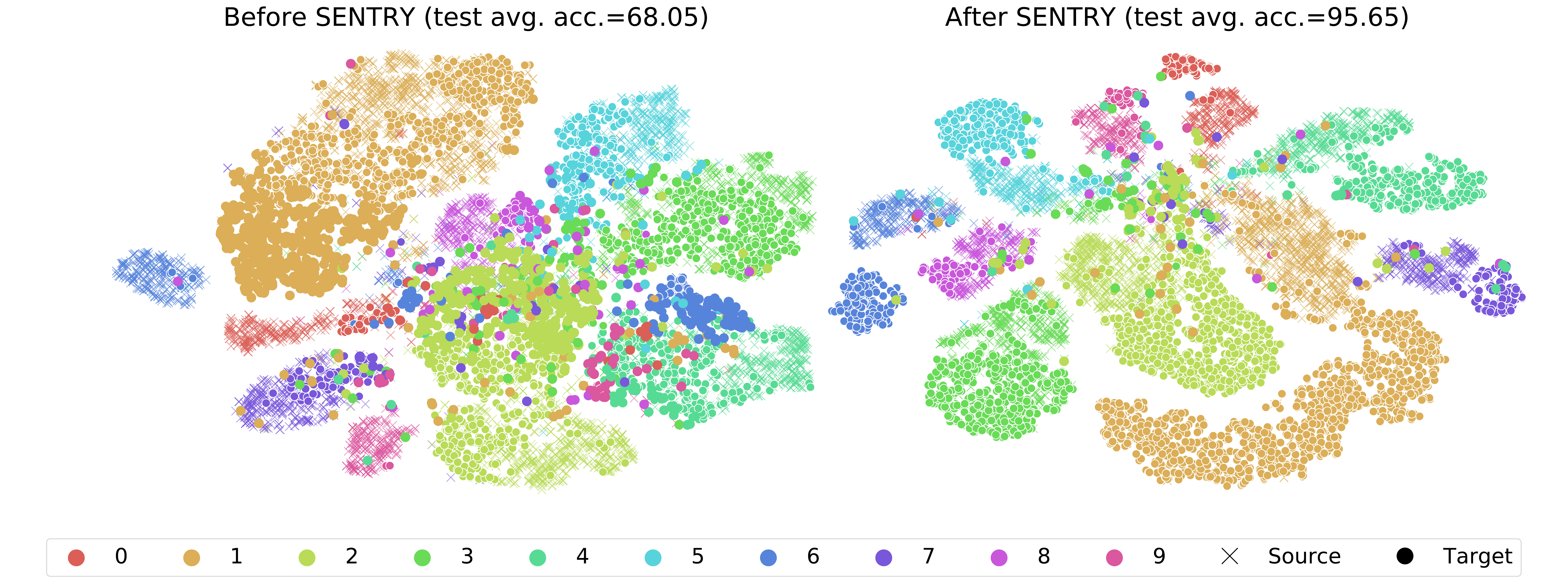}
    \caption[]{SVHN$\rightarrow$MNIST: We use t-SNE~\cite{maaten2008visualizing} to visualize features for incorrect (large, opaque circles) and correct (partly transparent circles) model predictions on the imbalanced target train set and source train set before (left) and after (right) adaptation via \method. Colors denote ground truth class, and $\times$ and $\bigcirc$ denote source and target instances. \method is able to overcome significant misalignments for both head classes with many examples (\eg 1's and 2's) as well as tail classes with very few examples (\eg 0's and 9's).}
    \label{fig:tsne}
\end{figure*}

\begin{figure*}[b]
    \centering
    \begin{subfigure}[b]{0.3\textwidth}  
        \centering 
        \includegraphics[width=\textwidth]{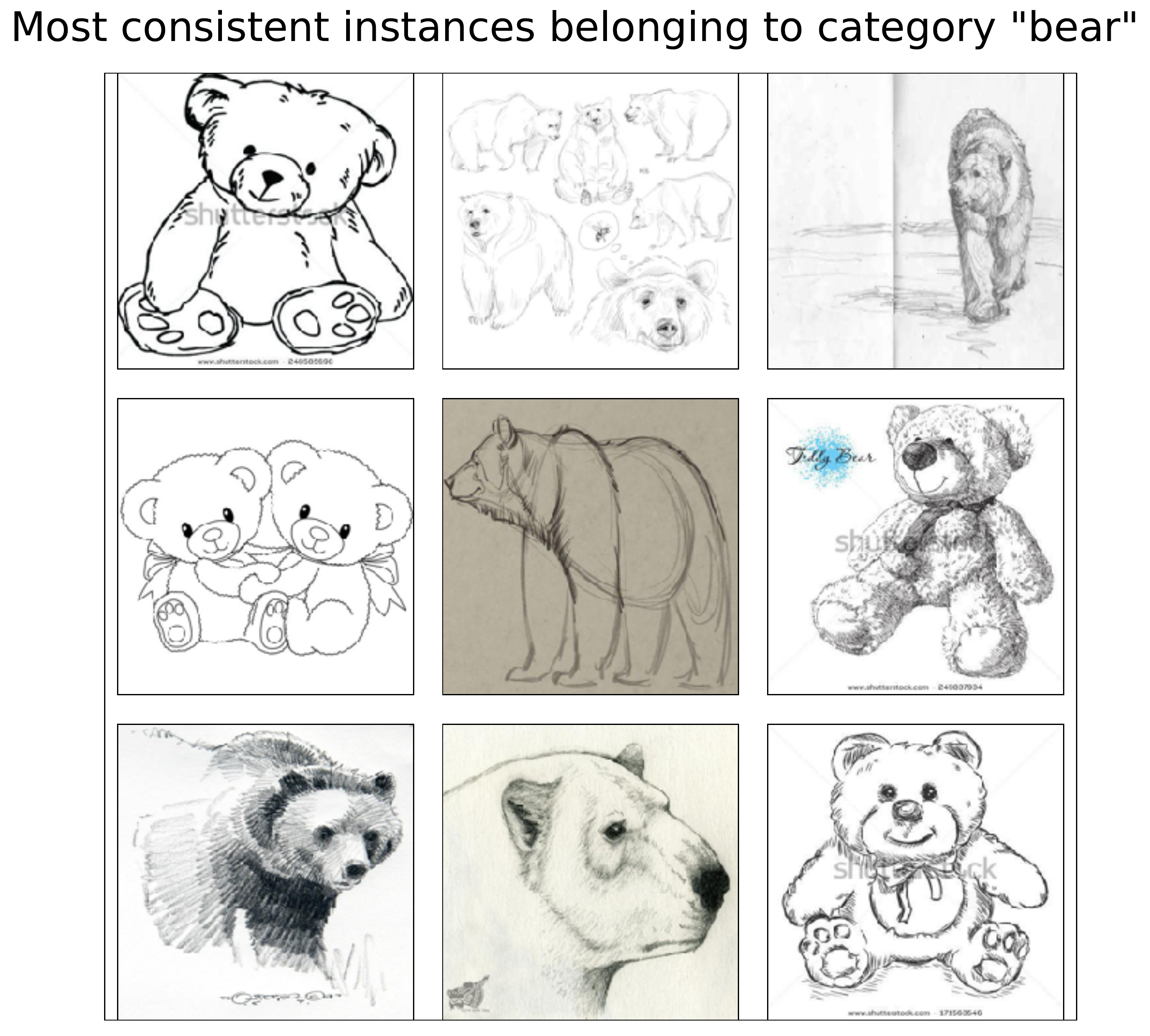}
        \caption[]%
        {{}}   
        \label{fig:consistent_bears}
    \end{subfigure}
    \begin{subfigure}[b]{0.3\textwidth}
        \centering
        \includegraphics[width=\textwidth]{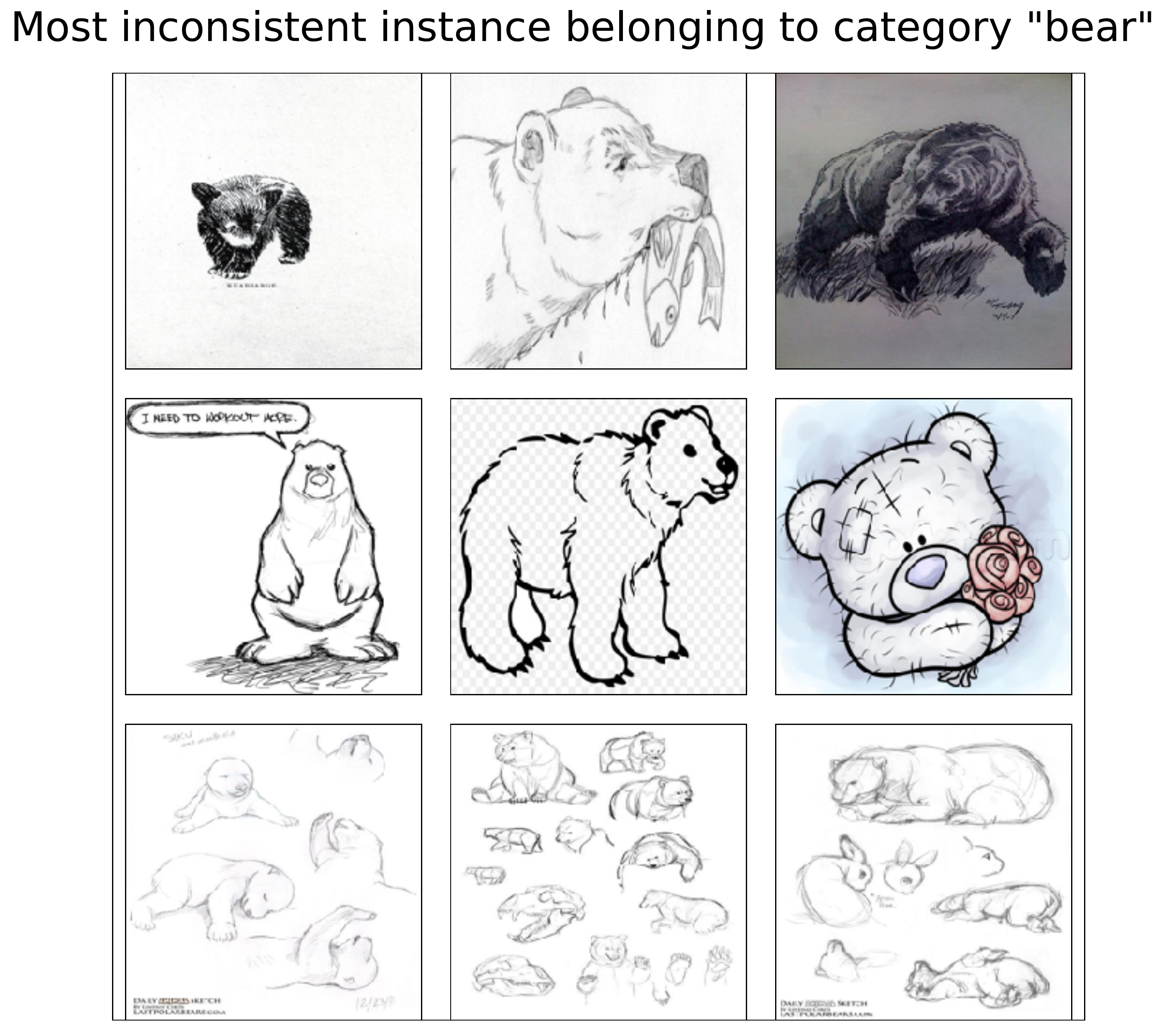}
        \caption[]%
        {{}}
        \label{fig:inconsistent_bears}
    \end{subfigure}
    \caption[]
    {DomainNet Clipart$\rightarrow$Sketch: Visualizing most consistent and inconsistent target instances.}
    \label{fig:qualitative_examples}
\end{figure*}

\noindent \textbf{t-SNE with \method.} Next, we use t-SNE~\cite{maaten2008visualizing} to visualize features (logits) extracted by the model for the source and target \emph{train} sets. In Fig.~\ref{fig:tsne}, we visualize the feature landscape before and after adaptation via \method. As seen, significant label imbalance exists: \eg a lot more 1's and 2's are present as compared to 0's and 9's. Further, we denote target instances that are \emph{incorrectly} classified by the model as large, opaque circles. Before adaptation (left), significant misalignments exist, particularly for head classes such as 1's and 2's. However, after adaptation via \method, cross-domain alignment for most classes improves significantly, as does the average accuracy on the target test set (68.1\% vs 95.7\%).

\noindent \textbf{Qualitative results.} In Fig~\ref{fig:qualitative_examples}, we provide some qualitative examples to build intuition about our consistency-based selection. For the Clipart$\rightarrow$Sketch shift, we visualize target (\ie sketch) instances belonging to the ground truth category ``bear''. On the left, we visualize a random subset of target instances for which model predictions are \emph{most consistent} under augmentations over the course of adaptation. On the right, we visualize a random subset of target instances for which model predictions are \emph{most inconsistent} over the course of adaptation via \method. Unsurprisingly, we find highly consistent instances to be ``easier'' to recognize, with more canonical poses and appearances. Similarly, inconsistent instances often tend to be challenging, and may even correspond to label noise, but our method appropriately avoids increasing model confidence on such instances.

\subsection{OfficeHome Results} 

In Table 3a of the main paper, we presented accuracies averaged over all 12 shifts in the standard version of OfficeHome~\cite{venkateswara2017deep} for our proposed method against prior work. In Table~\ref{tab:officehome_std}, we include the complete table with performances on every shift. As seen, \method achieves state-of-the-art performance on 9/12 shifts, and improves upon the next best method (InstaPBM~\cite{li2020rethinking}) by 2.5\% overall.

\begin{table*}[t]
    \begin{center} 
    \setlength{\tabcolsep}{2pt}
    \resizebox{\textwidth}{!}{
\begin{tabular}{lccccccccccccc}
    \toprule
    Method & \small{$\mathbf{Ar} \rightarrow \mathbf{Cl}$} & \small{$\mathbf{Ar} \rightarrow \mathbf{Pr}$} & \small{$\mathbf{Ar} \rightarrow \mathbf{Rw}$} & \small{$\mathbf{Cl} \rightarrow \mathbf{Ar}$} & \small{$\mathbf{Cl} \rightarrow \mathbf{Pr}$} & \small{$\mathbf{Cl} \rightarrow \mathbf{Rw}$} & \small{$\mathbf{Pr} \rightarrow \mathbf{Ar}$} & \small{$\mathbf{Pr} \rightarrow \mathbf{Cl}$} & \small{$\mathbf{Pr} \rightarrow \mathbf{Rw}$} & \small{$\mathbf{Rw} \rightarrow \mathbf{Ar}$} &  \small{$\mathbf{Rw} \rightarrow \mathbf{Cl}$} & \small{$\mathbf{Rw} \rightarrow \mathbf{Pr}$} & AVG \\
    \midrule
    Source & 34.9 & 50.0 & 58.0 & 37.4 & 41.9 & 46.2 & 38.5 & 31.2 & 60.4 & 53.9 & 41.2 & 59.9 & 46.1 \\
    \midrule
    DAN~\cite{long2015learning} & 43.6 & 57.0 & 67.9 & 45.8 & 56.5 & 60.4 & 44.0 & 43.6 & 67.7 & 63.1 & 51.5 & 74.3 & 56.3 \\
    DANN~\cite{ganin2014unsupervised} & 45.6 & 59.3 & 70.1 & 47.0 & 58.5 & 60.9 & 46.1 & 43.7 & 68.5 & 63.2 & 51.8 & 76.8 & 57.6 \\
    JAN~\cite{long2017deep} & 45.9 & 61.2 & 68.9 & 50.4 & 59.7 & 61.0 & 45.8 & 43.4 & 70.3 & 63.9 & 52.4 & 76.8 & 58.3 \\
    CDAN~\cite{long2018conditional} & 50.7 & 70.6 & 76.0 & 57.6 & 70.0 & 70.0 & 57.4 & 50.9 & 77.3 & 70.9 & 56.7 & 81.6 & 65.8 \\
    BSP~\cite{chen2019transferability} & 52.0 & 68.6 & 76.1 & 58.0 & 70.3 & 70.2 & 58.6 & 50.2 & 77.6 & 72.2 & 59.3 & 81.9 & 66.3\\
    MDD~\cite{zhang2019bridging} & 54.9 & 73.7 & 77.8 & 60.0 & 71.4 & 71.8 & 61.2 & 53.6 & 78.1 & 72.5 & 60.2 & 82.3 & 68.1 \\
    MCS & 55.9 & 73.8 & 79.0 & 57.5 & 69.9 & 71.3 & 58.4 & 50.3 & 78.2 & 65.9 & 53.2 & 82.2 & 66.3 \\      
    InstaPBM~\cite{li2020rethinking} & 54.4 & 75.3 & 79.3 & \underline{65.4} & \textbf{74.2} & \textbf{75.0} & 63.3 & 49.7 & \underline{80.2} & \underline{72.8} & 57.8 & \underline{83.6} & \underline{69.7} \\              
    MDD+I.A~\cite{jiang2020implicit} & \underline{56.2} & \textbf{77.9} & \underline{79.2} & 64.4 & \underline{73.1} & 74.4 & \underline{64.2} & \underline{54.2} & 79.9 & 71.2 & \underline{58.1} & 83.1 & 69.5 \\    
    \midrule
    Ours & \textbf{61.8} & \underline{77.4} & \textbf{80.1} & \textbf{66.3} & 71.6 & \underline{74.7} & \textbf{66.8} & \textbf{63.0} & \textbf{80.9} & \textbf{74.0} & \textbf{66.3} & \textbf{84.1} & \textbf{72.2} \\
    \bottomrule
    \end{tabular}}
    \caption{Accuracies on standard OfficeHome. Bold and underscore denote the best and second-best performing methods respectively.}
    \label{tab:officehome_std}
    \end{center}
\end{table*}

\begin{figure*}[t]
    \centering
    \includegraphics[width=0.9\textwidth]{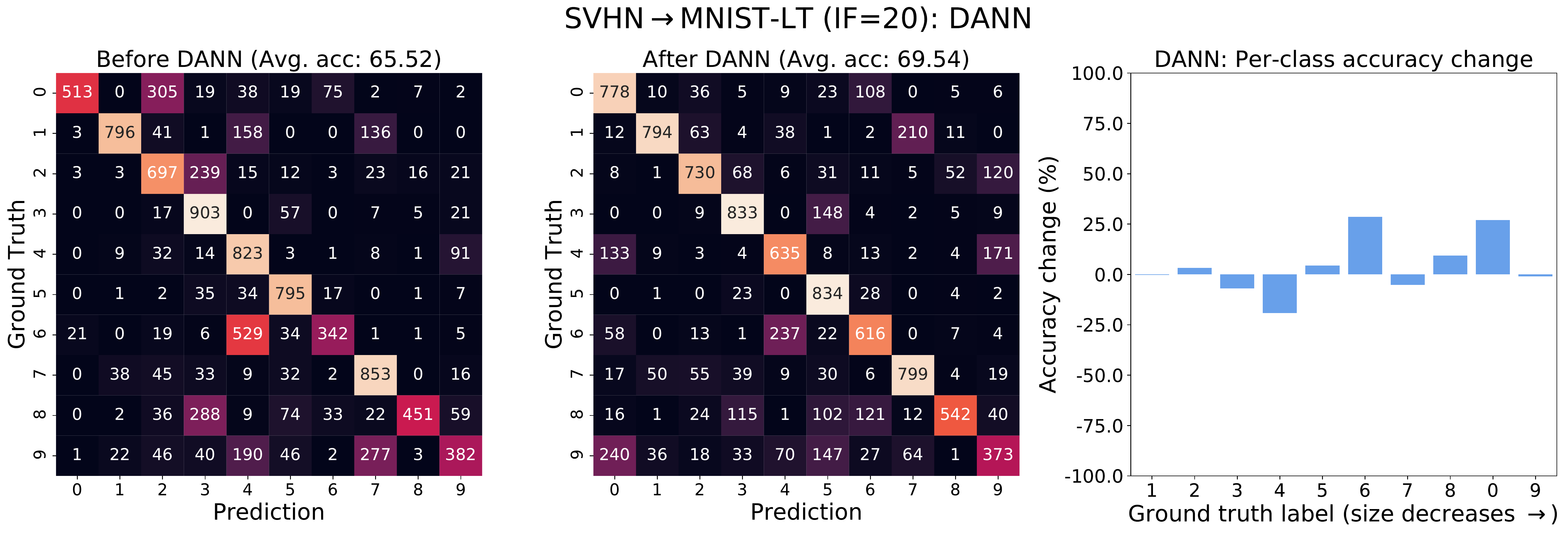}
    \caption{
        SVHN$\rightarrow$MNIST-LT (IF=20): Performance on target test set after \texttt{DANN}~\cite{ganin2014unsupervised}.
    }
    \label{fig:dann_cm}
 \end{figure*}

\subsection{Additional Implementation Details} 

In Sec 4.2 of the main paper, we presented implementation details for our method. We describe a few additional details to aid in reproducibility.

\noindent \textbf{Training and Optimization.} We match optimization details to Tan~\etal~\cite{tan2019generalized}. On all benchmarks other than DIGITS, we use SGD with momentum of 0.9, a learning rate of $10^{-2}$ for the last layer and $10^{-3}$ for all other layers, and weight decay of $5\times10^{-4}$. We use the learning rate decay strategy proposed in Ganin~\etal~\cite{ganin2014unsupervised}. On DIGITS, we use Adam with a learning rate of $2\times10^{-4}$ and no weight decay. We use a batch size of 16 on DomainNet, OfficeHome, and VisDA, and 128 on DIGITS. For data augmentation when training the source models on DomainNet, OfficeHome, and VisDA, we first resize to 256 pixels, extract a random crop of size (224x224), and randomly flip images with a 50\% probability. For \method, we use RandAugment~\cite{cubuk2020randaugment} for generating augmented images, as described in Sec. 3.3 of the main paper and do not use any additional augmentations. For $L_{\method}$, we average loss for consistent and inconsistent instances separately and weigh each loss by the proportion of instances assigned to each group. We select $\lambda_{IE}$=0.1 and $\lambda_{\method}$=1.0 so as to approximately scale each loss term to the same order of magnitude.

\noindent \textbf{Baseline implementations}. For all baselines except InstaPBM~\cite{li2020rethinking}, we directly report results from prior work. We base our InstaPBM implementation on code provided by authors and implement target information entropy, conditional entropy, contrastive, and mixup losses with loss weights 0.1, 1.0, 0.01, 0.1 respectively.

\begin{figure*}
    \centering
    \begin{subfigure}[b]{0.24\textwidth}  
        \centering 
        \includegraphics[width=\textwidth]{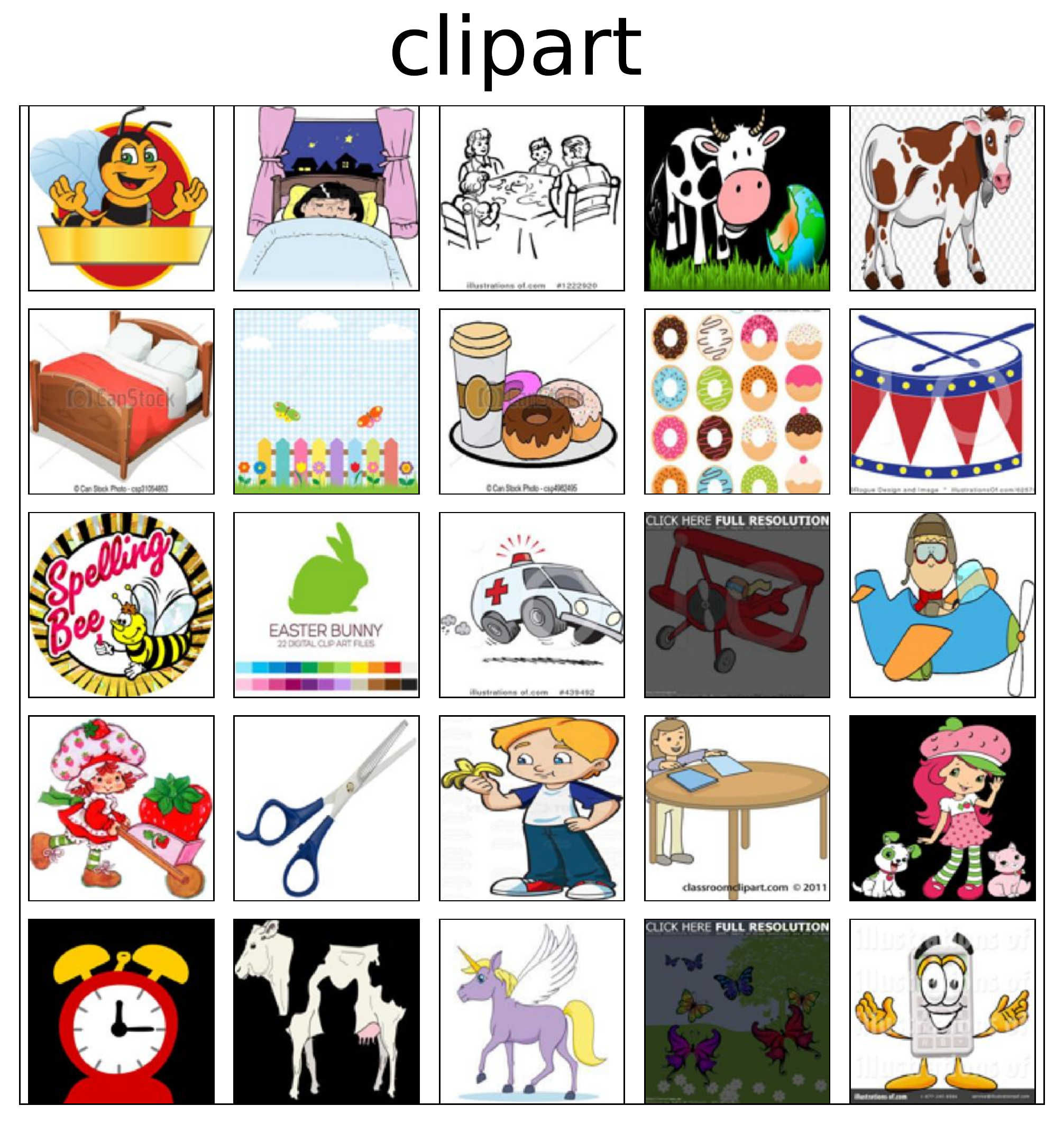}
        \caption[]%
        {{}}   
        \label{fig:clipart}
    \end{subfigure}
    \begin{subfigure}[b]{0.24\textwidth}
        \centering
        \includegraphics[width=\textwidth]{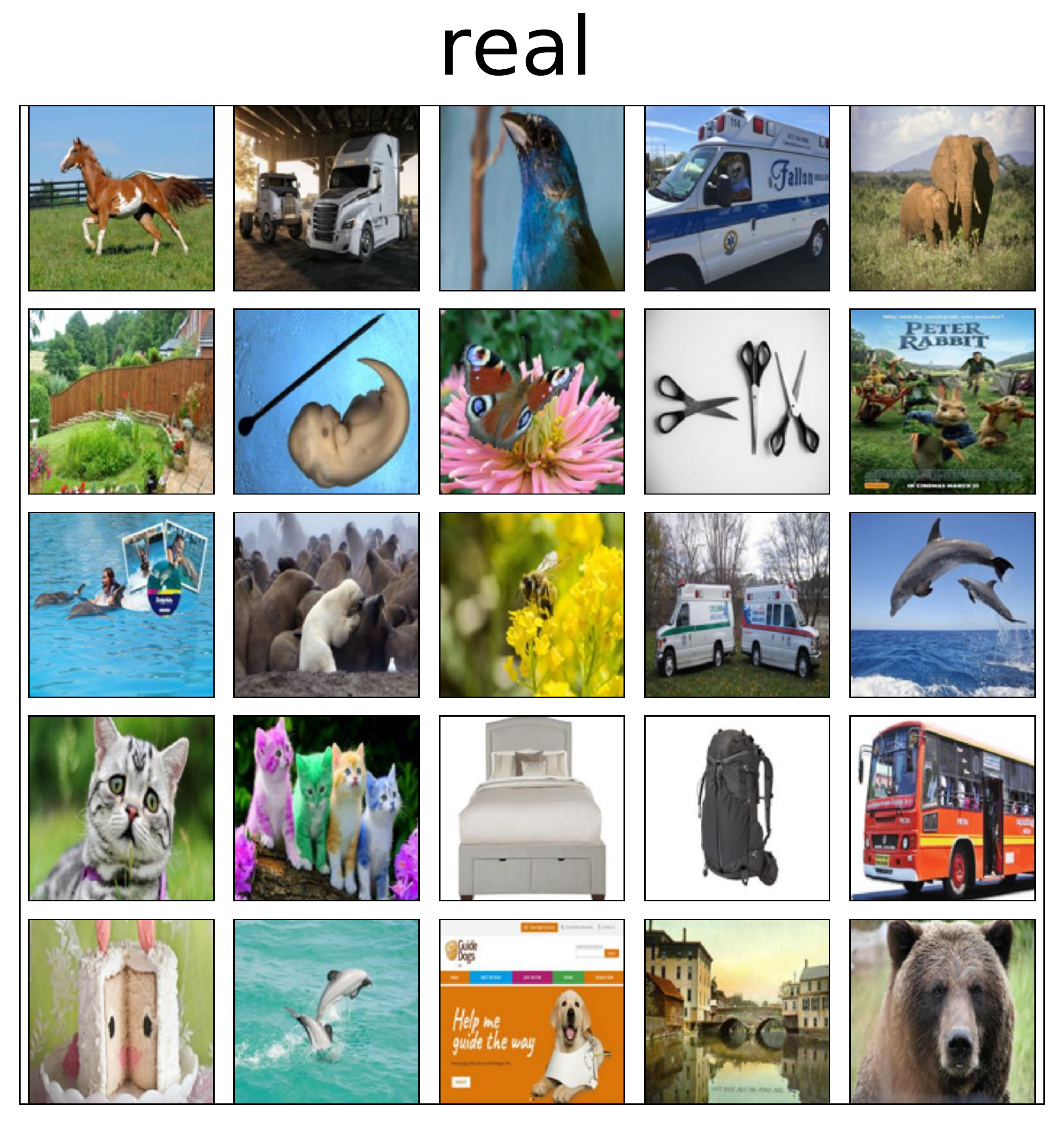}
        \caption[]%
        {{}}
        \label{fig:real}
    \end{subfigure}
    \begin{subfigure}[b]{0.24\textwidth}
        \centering
        \includegraphics[width=\textwidth]{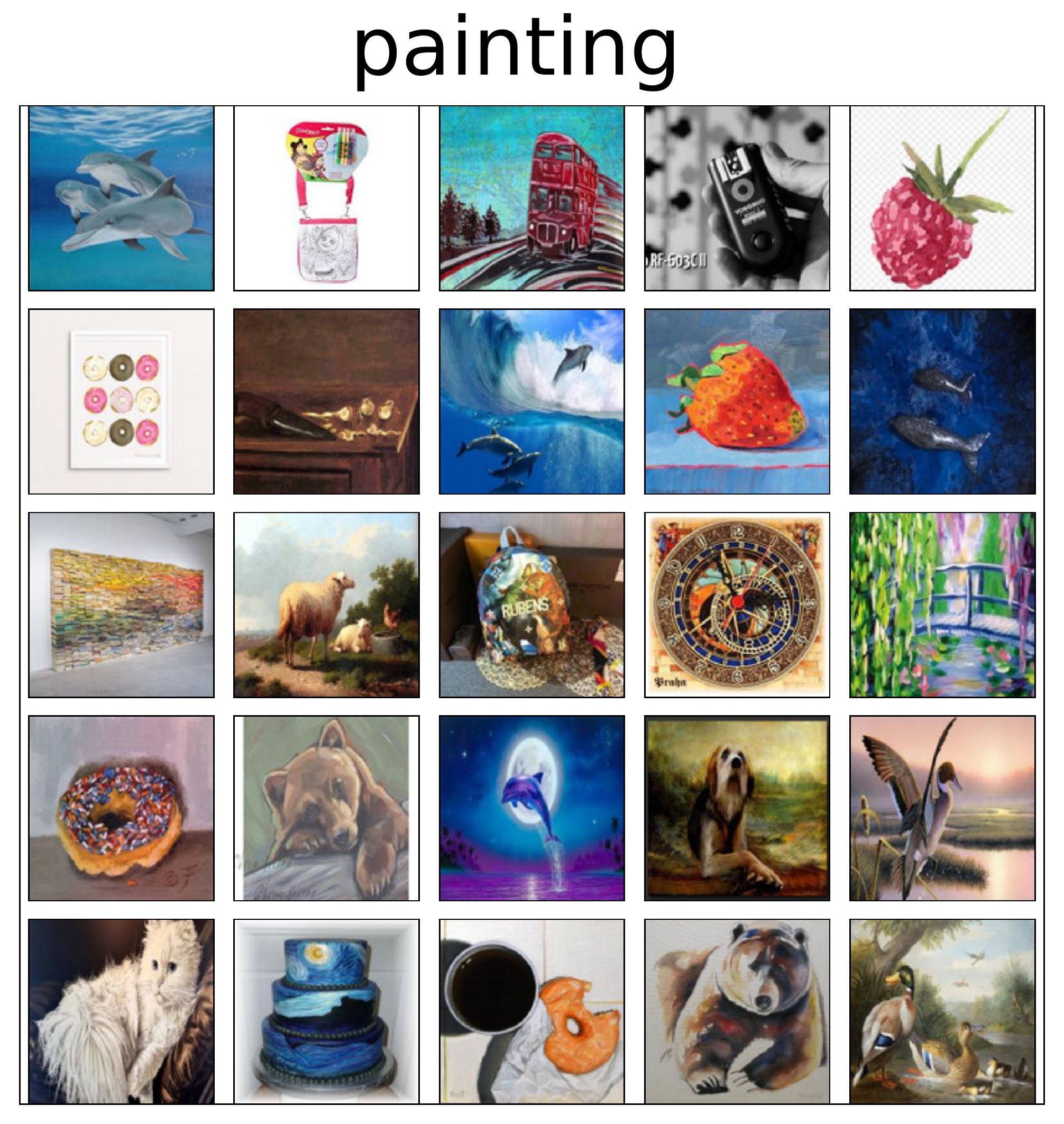}
        \caption[]%
        {{}}
        \label{fig:painting}
    \end{subfigure}
    \begin{subfigure}[b]{0.24\textwidth}
        \centering
        \includegraphics[width=\textwidth]{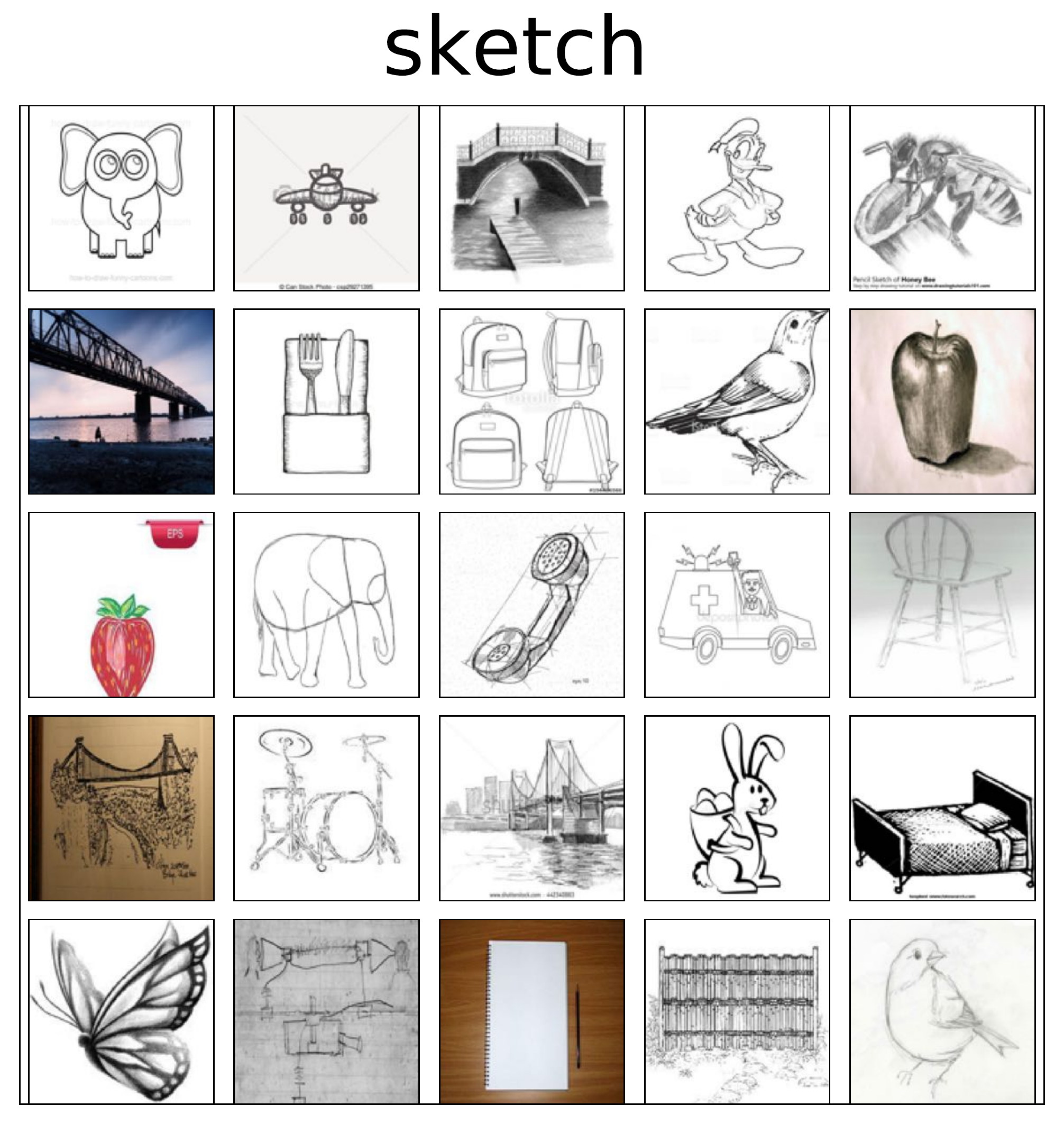}
        \caption[]%
        {{}}
        \label{fig:sketch}
    \end{subfigure}
    \begin{subfigure}[b]{0.7\textwidth}
        \centering
        \includegraphics[width=\textwidth]{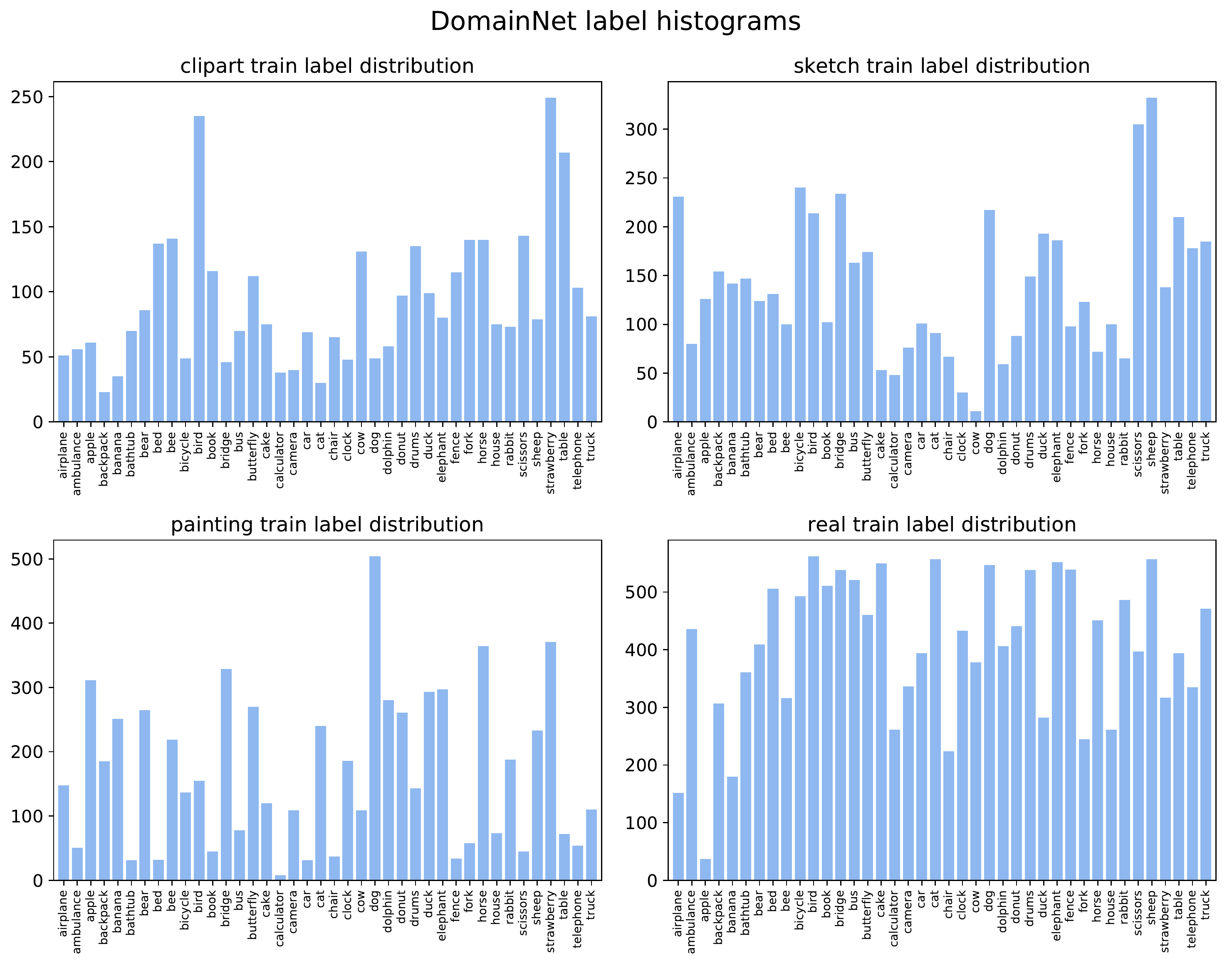}
        \caption[]%
        {{}}
        \label{fig:minidn}
    \end{subfigure}
    \caption[]
    {DomainNet~\cite{peng2019moment} statistics: (a)-(d): Qualitative examples from each domain. (e): Label histograms for the splits proposed in Tan~\etal~\cite{tan2019generalized}.}
    \label{fig:domainnet_qual}
\end{figure*}

\subsection{Analysis of DM-based methods under LDS} 

Prior work has already demonstrated the shortcomings of distribution-matching based UDA methods under additional label distribution shift~\cite{wu2019domain,li2020rethinking}. Wu~\etal~\cite{wu2019domain} show that DM-based domain adversarial methods optimize two out of a sum of three terms that bound target error as shown in Ben-David~\etal~\cite{ben2010theory}. Under matching task label distributions across domains, the contribution of the third term is small, which is the assumption under which these methods operate; absent this, the third term is unbounded and DM-based methods are not expected to succeed in domain alignment. We refer readers to Sec 2 of their paper for a formal proof. 

Under LDS, such DM-based methods are expected to primarily mis-align majority (head) classes in the target domain with other classes in the source domain. We empirically test this hypothesis on the SVHN$\rightarrow$MNIST-LT (IF=20) domain shift for digit recognition. In Fig~\ref{fig:dann_cm}, we repeat our per-class accuracy analysis for adaptation via \texttt{DANN}~\cite{ganin2014unsupervised}, a popular distribution matching UDA algorithm that uses domain adversarial feature matching. \texttt{DANN} has been shown to lead to successful domain alignment in the \emph{absence} of label distribution shift (LDS); we now test its effectiveness in the presence of LDS. As seen, significant misalignments exist before adaptation (left). To match the source training strategy and architecture to the original paper, we do not use the few-shot architecture we use for our method, which leads to the slightly lower starting performance observed as compared to Fig.~\ref{fig:sentry_cm}. However, due to label imbalance, \texttt{DANN} is unable to appropriately align instances and only slightly improves performance (69.54\% in Fig~\ref{fig:dann_cm}, middle). In Fig~\ref{fig:dann_cm}, we show the \emph{change} in accuracy for each class after adaptation, while sorting classes in decreasing order of size. As predicted by the theory, \texttt{DANN} does particularly poorly on head (majority) classes (1, 2, 3, 4), while slightly improving performance for classes with fewer examples. This is in contrast to our method \method, which is able to improve performance for both head and tail classes (Fig~\ref{fig:sentry_cm}, right).

\subsection{Dataset Details} 

 \begin{figure*}
    \centering   
    \begin{subfigure}[b]{0.24\textwidth}
        \centering
        \includegraphics[width=\textwidth]{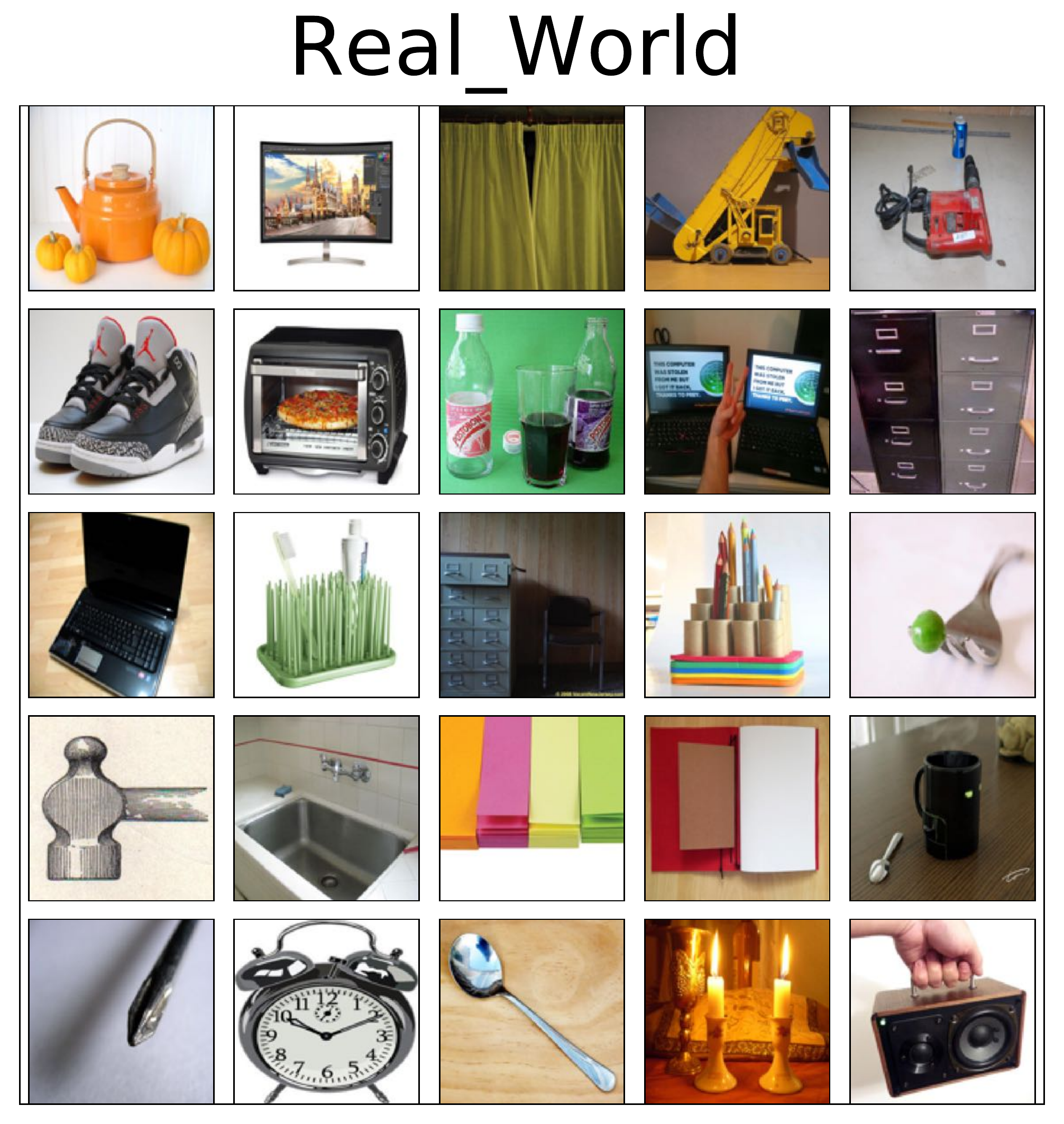}
        \caption[]%
        {{}}
        \label{fig:real_world}
    \end{subfigure}
    \begin{subfigure}[b]{0.24\textwidth}
        \centering
        \includegraphics[width=\textwidth]{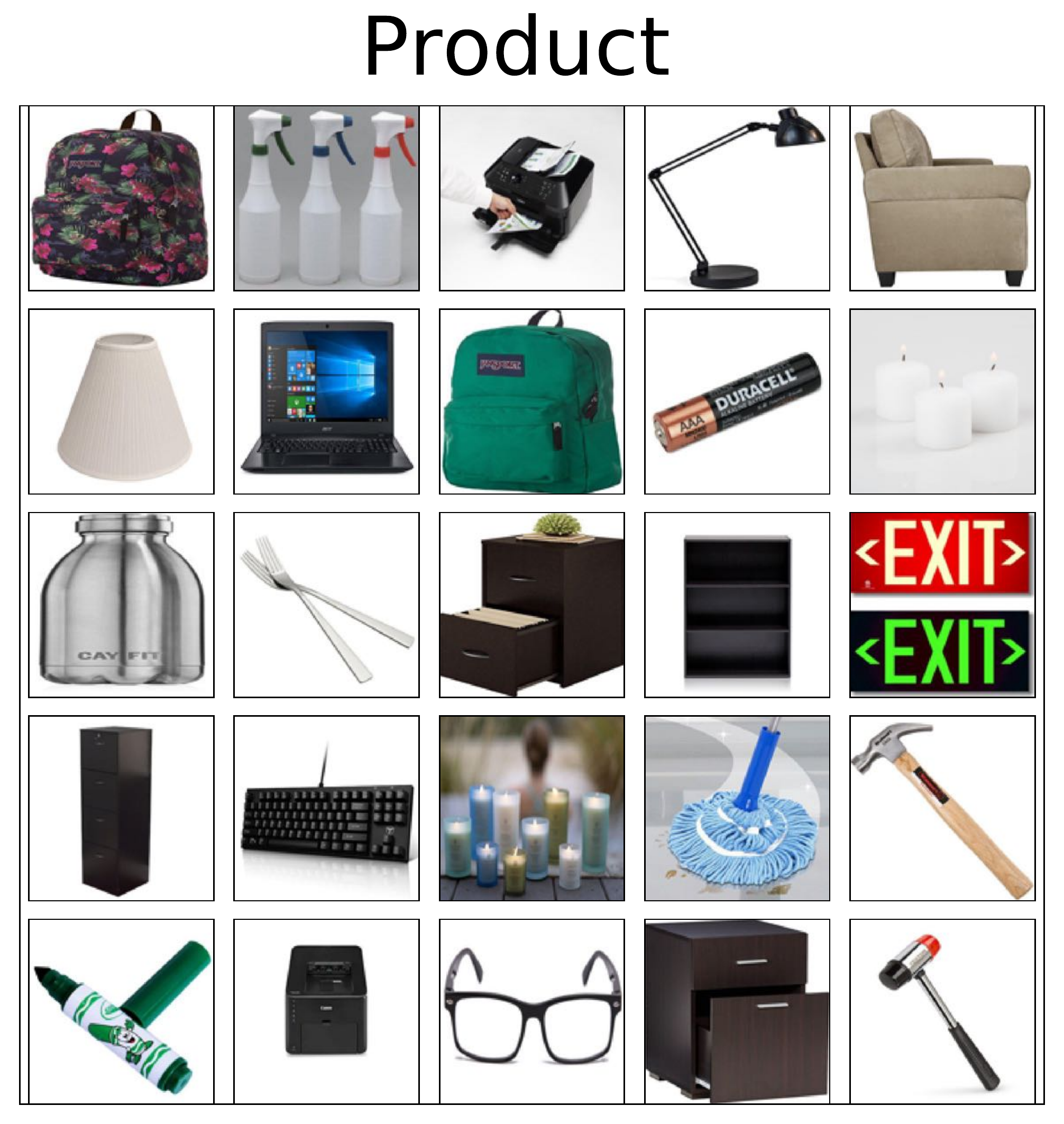}
        \caption[]%
        {{}}
        \label{fig:product}
    \end{subfigure}
    \begin{subfigure}[b]{0.24\textwidth}
        \centering
        \includegraphics[width=\textwidth]{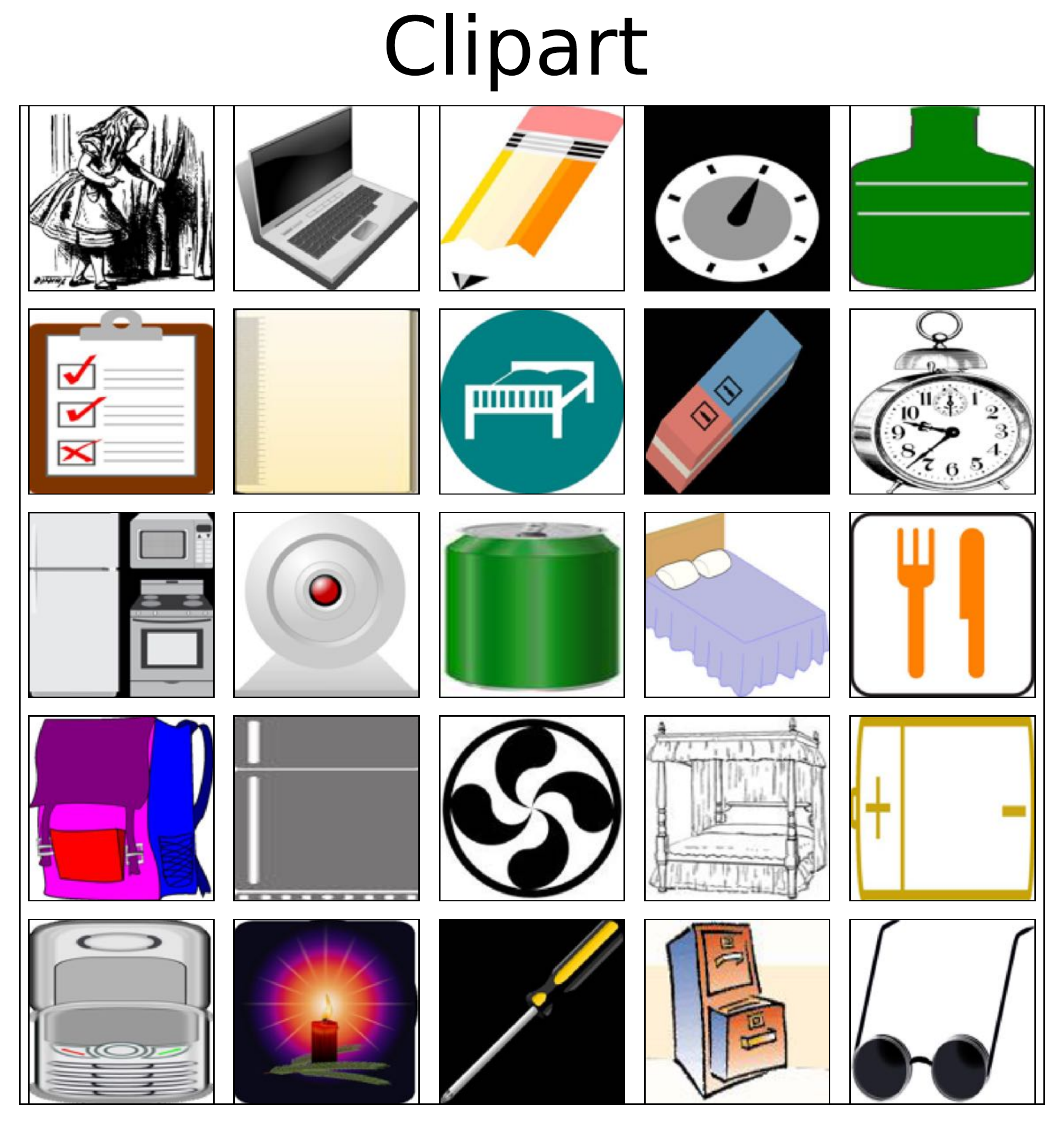}
        \caption[]%
        {{}}
        \label{fig:clipart_OH}
    \end{subfigure}
    \begin{subfigure}[b]{\textwidth}
        \centering
        \includegraphics[width=\textwidth]{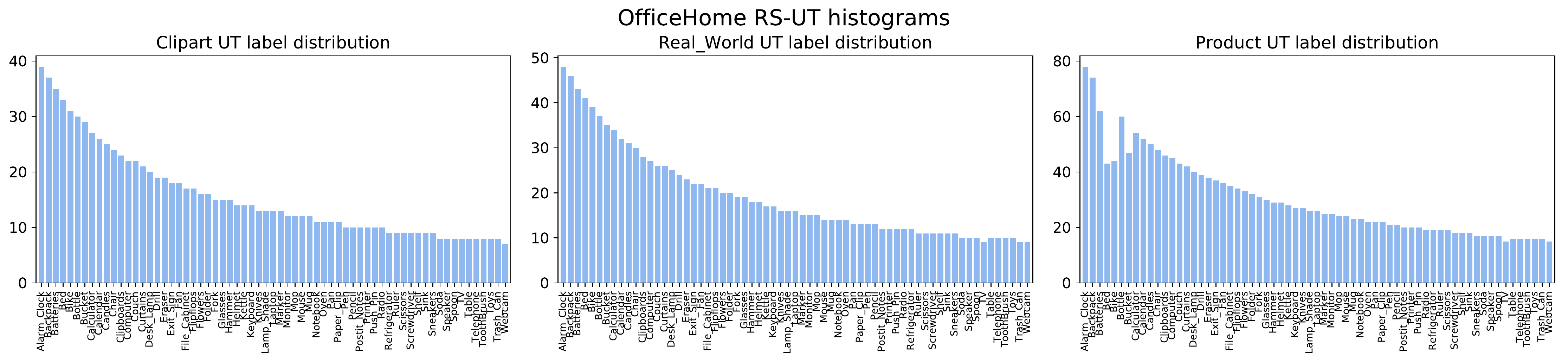}
        \caption[]%
        {{}}
        \label{fig:ohrsut}
    \end{subfigure}
    \caption[]
    {OfficeHome~\cite{venkateswara2017deep} statistics: (a)-(d): Qualitative examples from each domain. (e): Label histograms for the UT target splits proposed in Tan~\etal~\cite{tan2019generalized}.}
    \label{fig:ohrsut_qual}
\end{figure*}

In Sec 4.1, we described our datasets in detail. For completeness, we also include label histograms and qualitative examples from each domain in the DomainNet and OfficeHome RS-UT benchmarks proposed in Tan~\etal~\cite{tan2019generalized} in Figs.~\ref{fig:domainnet_qual},~\ref{fig:ohrsut_qual}.

\par \noindent \textbf{Idiosyncrasy of the ``clipart'' domain}. Some prior works in UDA (\eg Tan~\etal~\cite{tan2019generalized}) use center cropping at test time on DomainNet and OfficeHome, wherein they first resize a given image to 256 pixels and then extract a 224x224 crop from the center of the image. This practice has presumably carried over from ImageNet evaluation, where images are known to have a center bias~\cite{torralba2011unbiased}. Figs.~\ref{fig:clipart},~\ref{fig:clipart_OH} show qualitative examples from the clipart domain in DomainNet and OfficeHome. As seen, most clipart categories span the entire extent of the image and do \emph{not} have a center bias. As a result, using center cropping at evaluation time can adversely affect performance when adapting to Clipart as a target domain. We show empirical evidence of this in Tables~\ref{tab:domainnet_ccrop},~\ref{tab:officehome_ccrop} -- when clipart is the target domain, performance drops consistently when using centercrop at test time. For \method, we therefore do not use center crop at evaluation time. %
For comparison, we also include the performance of our strongest baseline in each setting in Tabs.~\ref{tab:domainnet_ccrop},~\ref{tab:officehome_ccrop}: InstaPBM~\cite{li2020rethinking} and MDD+I.A.~\cite{jiang2020implicit}, respectively. However, we note that in both settings, with and without centercrop, \method still clearly outperforms our strongest baselines on both benchmarks.

\begin{table*}[t]
    \begin{center} 
    \setlength{\tabcolsep}{2pt}
    \resizebox{\textwidth}{!}{
    \begin{tabular}{l c c c c c c c c c c c c c}
        \toprule
        Method & $\mathbf{R} \rightarrow \mathbf{C}$ & $\mathbf{R} \rightarrow \mathbf{P}$ & $\mathbf{R} \rightarrow \mathbf{S}$ & $\mathbf{C} \rightarrow \mathbf{R}$ & $\mathbf{C} \rightarrow \mathbf{P}$ & $\mathbf{C} \rightarrow \mathbf{S}$ & $\mathbf{P} \rightarrow \mathbf{R}$ & $\mathbf{P} \rightarrow \mathbf{C}$ & $\mathbf{P} \rightarrow \mathbf{S}$ & $\mathbf{S} \rightarrow \mathbf{R}$ & $\mathbf{S} \rightarrow \mathbf{C}$ & $\mathbf{S} \rightarrow \mathbf{P}$ & AVG \\
        \midrule
        source & 65.75 & 68.84 & 59.15 & 77.71 & 60.60 & 57.87 & 84.45 & 62.35 & 65.07 & 77.10 & 63.00 & 59.72 & 66.80 \\
        \rowcolor{Gray}
        \texttt{+CC-eval} & \textcolor{red}{62.42} & 69.31 & 59.28 & 79.74 & 59.49 & 58.46 & 84.55 & \textcolor{red}{60.42} & 66.26 & 78.61 & \textcolor{red}{58.31} & 61.31 & 66.51 \\
        \midrule     
        InstaPBM~\cite{li2020rethinking} & 80.10 & 75.87 & 70.84 & 89.67 & 70.21 & 72.76 & 89.60 & 74.41 & 72.19 & 87.00 & 79.66 & 71.75 & 77.84 \\
        \midrule
        \method (Ours) & 83.89 & 76.72 & 74.43 & 90.61 & 76.02 & 79.47 & 90.27 & 82.91 & 75.60 & 90.41 & 82.40 & 73.98 & 81.39 \\        
        \rowcolor{Gray}
        \texttt{+CC-eval} & \textcolor{red}{78.81} & 78.15 & 71.62 & 89.84 & 75.98 & 77.69 & 89.50 & \textcolor{red}{77.34} & 73.82 & 89.96 & \textcolor{red}{80.66} & 75.02 & 79.87 \\
        \bottomrule
        \end{tabular}}
        \vspace{-5pt}
        \caption{Idiosyncrasy of the ``clipart'' domain: Per-class average accuracies on DomainNet without (white rows) and with (gray rows) centercrop at test time. We highlight in red performance drops due to centercrop eval when adaptating to Clipart as a target domain. For comparison, we also include the performance of InstaPBM~\cite{li2020rethinking}, the 2nd best method from Table 1 in the main paper.}\label{tab:domainnet_ccrop}
        \vspace{-5pt}
    \end{center}
    \end{table*}

    \begin{table}[t]
        \begin{center} 
        \setlength{\tabcolsep}{1.5pt}
        \resizebox{\columnwidth}{!}{
    \begin{tabular}{lccccccc}
        \toprule        
        Method & \small{$\mathbf{Rw}\veryshortarrow\mathbf{Pr}$} & \small{$\mathbf{Rw}\veryshortarrow\mathbf{Cl}$} & \small{$\mathbf{Pr}\veryshortarrow\mathbf{R w}$} & \small{$\mathbf{Pr}\veryshortarrow\mathbf{Cl}$} & \small{$\mathbf{Cl}\veryshortarrow\mathbf{R} \mathbf{w}$} & \small{$\mathbf{Cl}\veryshortarrow\mathbf{Pr}$} & AVG \\
        \midrule
        source & 70.74 & 44.24 & 67.33 & 38.68 & 53.51 & 51.85 & 54.39 \\
        \rowcolor{Gray}
        \texttt{+CC-eval} & 70.25 & \textcolor{red}{38.20} & 67.74 & \textcolor{red}{35.61} & 55.08 & 52.90 & 53.30 \\
        \midrule
        MDD+I.A~\cite{jiang2020implicit} & 76.08 & 50.04 & 74.21 & 45.38 & 61.15 & 63.15 & 61.67 \\
        \midrule
        \method (Ours) & 76.12 & 56.80 & 73.60 & 54.75 & 65.94 & 64.29 & 65.25 \\        
        \rowcolor{Gray}
        \texttt{+CC-eval} & 76.35 & \textcolor{red}{52.25} & 73.08 & \textcolor{red}{50.60} & 66.69 & 64.19 & 63.86 \\
        \bottomrule
        \end{tabular}
        }
        \caption{Idiosyncrasy of the ``clipart'' domain: Per-class average accuracies on OfficeHome RS-UT without (white rows) and with (gray rows) centercrop at test time. We highlight in red performance drops due to centercrop eval when adaptating to Clipart as target. For comparison, we also include the performance of MDD+I.A.~\cite{jiang2020implicit}, the 2nd best method from Table 2 in the main paper.}
        \vspace{-10pt}
        \label{tab:officehome_ccrop}
        \end{center}
    \end{table}

\end{document}